\pdfoutput=1 

\documentclass[letterpaper]{article}

\usepackage{microtype}
\usepackage{graphicx}
\usepackage{subfigure}
\usepackage{booktabs} %

\usepackage{framed}
\usepackage{hyperref}
\usepackage{breakurl}
\hypersetup{
           breaklinks=true,   %
           colorlinks=true,   %
           pdfusetitle=true,  %
        }
        
\usepackage{scalefnt}

\usepackage[accepted]{icml2024}

\usepackage{amsmath}
\usepackage{amssymb}
\usepackage{mathtools}
\usepackage{amsthm}

\usepackage[capitalize,noabbrev]{cleveref}

\usepackage{microtype}
\usepackage{inconsolata}
\usepackage{graphicx}
\usepackage{subfigure}
\usepackage{booktabs} %
\usepackage{amssymb}
\usepackage{bm}
\usepackage{bbm}
\usepackage{amsmath}  %
\usepackage{amsthm}
\usepackage{xcolor, color}
\usepackage{textcomp}
\usepackage{multirow}
\usepackage{mathtools}
\usepackage{algorithm}
\usepackage{algpseudocode}
\usepackage{listings,multicol}  %

\usepackage{rotating} 

\algrenewcommand\algorithmicrequire{\textbf{Input:}}
\algrenewcommand\algorithmicensure{\textbf{Output:}}

\usepackage{xspace}
\usepackage{pifont}
\newcommand{\xmark}{\ding{55}}
\usepackage{wrapfig}
\usepackage{tikz}
\usepackage{pifont}
\usepackage{pgfplots}
\usepackage[labelfont=bf,font=small]{caption}
\usepackage{wasysym}

\usetikzlibrary{arrows}
\usetikzlibrary{positioning}
\tikzset{
  treenode/.style = {align=center, inner sep=0pt, text centered,
    font=\sffamily},
  arn_n/.style = {treenode, circle, black, font=\sffamily\bfseries, draw=black,
    fill=white, text width=1.5em},%
  arn_r/.style = {treenode, circle, black, font=\sffamily\bfseries, draw=black,
    fill=white, text width=1.0em},%
  arn_x/.style = {treenode, rectangle, draw=black,
    minimum width=0.5em, minimum height=0.5em}%
}

\definecolor{tabblue}{HTML}{4e79a7}
\definecolor{tabred}{HTML}{e15759}
 \usepackage{url}

\usepackage{bm}
\usepackage{enumitem}

\usepackage{subcaption}
\usepackage{fontawesome}

\usepackage[compact]{titlesec}
\usepackage{etoolbox}
\newcommand{\zerodisplayskips}{%
  \setlength{\abovedisplayskip}{4pt}%
  \setlength{\belowdisplayskip}{4pt}%
  \setlength{\abovedisplayshortskip}{4pt}%
  \setlength{\belowdisplayshortskip}{4pt}}
\appto{\normalsize}{\zerodisplayskips}
\appto{\small}{\zerodisplayskips}
\appto{\footnotesize}{\zerodisplayskips}
\usepackage[subtle, mathdisplays=normal, charwidths=normal, leading=normal]{savetrees}

\def\balpha{{\bm{\alpha}}}
\def\bbeta{{\bm{\beta}}}
\def\bgamma{{\bm{\gamma}}}

\def\bLambda{{\mathbf{\Lambda}}}
\def\bGamma{{\mathbf{\Gamma}}}

\providecommand{\lnorm}{\operatorname{LN}}

\providecommand{\swiglu}{\text{SwiGLU}}

\providecommand{\revcum}{\text{revcum}}

\algblock{ParFor}{EndParFor}
\algnewcommand\algorithmicparfor{\textbf{parfor}}
\algnewcommand\algorithmicpardo{\textbf{do}}
\algnewcommand\algorithmicendparfor{\textbf{end\ parfor}}
\algrenewtext{ParFor}[1]{\algorithmicparfor\ #1\ \algorithmicpardo}
\algrenewtext{EndParFor}{\algorithmicendparfor}

\def\rmdO{{\mathbf{dO}}}
\def\rmdQ{{\mathbf{dQ}}}
\def\rmdK{{\mathbf{dK}}}
\def\rmdtQ{{\mathbf{d\tilde Q}}}
\def\rmdtK{{\mathbf{d \tilde K}}}
\def\rmdbK{{\mathbf{d \bar K}}}
\def\rmdP{{\mathbf{dP}}}
\def\rmdS{{\mathbf{dS}}}

\def\rmdG{{\mathbf{dG}}}

\def\rmdA{{\mathbf{dA}}}

\def\rmdV{{\mathbf{dV}}}

\def\vlogb{{\log\bm{b}}}
\def\vlogd{{\log\bm{b=d}}}
\def\vlogalpha{{\log\bm{\alpha}}}
\def\vdlogb{{\mathbf{d}\log\bm{b}}}
\def\vdlogd{{\mathbf{d}\log\bm{d}}}
\def\vdk{{\mathbf{d}\bm{k}}}

\def\vdq{{\mathbf{d}\bm{q}}}
\def\vdv{{\mathbf{d}\bm{v}}}
\def\vdo{{\mathbf{d}\bm{o}}}
\def\dbalpha{{\mathbf{d}\bm{\alpha}}}
\def\dbbeta{{\mathbf{d}\bm{\beta}}}
\def\dblogalpha{{\mathbf{d}\log\bm{\alpha}}}
\def\dblogbeta{{\mathbf{d}\log\bm{\beta}}}

\definecolor{bred}{RGB}{250, 82, 82}
\definecolor{borange}{RGB}{253, 126, 20}
\definecolor{byellow}{RGB}{250, 176, 5}
\definecolor{bgreen}{RGB}{116, 184, 22}
\definecolor{bblue}{RGB}{250, 176, 5}
\definecolor{bindigo}{RGB}{76, 110, 245}
\definecolor{bcyan}{RGB}{59, 201, 219}
\definecolor{bteal}{RGB}{99, 230, 190}

\usepackage{amsmath,amsfonts,bm}

\renewcommand\intercal{{\cramped{{}^\mathsf{T}}}}
\def\eqref#1{equation~\ref{#1}}

\def\1{\bm{1}}

\def\rmA{{\mathbf{A}}}
\def\rmB{{\mathbf{B}}}

\def\rmD{{\mathbf{D}}}

\def\rmG{{\mathbf{G}}}

\def\rmK{{\mathbf{K}}}

\def\rmM{{\mathbf{M}}}

\def\rmO{{\mathbf{O}}}
\def\rmP{{\mathbf{P}}}
\def\rmQ{{\mathbf{Q}}}

\def\rmS{{\mathbf{S}}}

\def\rmV{{\mathbf{V}}}

\def\rmX{{\mathbf{X}}}
\def\rmY{{\mathbf{Y}}}
\def\rmZ{{\mathbf{Z}}}

\def\vzero{{\bm{0}}}

\def\va{{\bm{a}}}
\def\vb{{\bm{b}}}
\def\vc{{\bm{c}}}
\def\vd{{\bm{d}}}

\def\vk{{\bm{k}}}

\def\vo{{\bm{o}}}

\def\vq{{\bm{q}}}
\def\vr{{\bm{r}}}

\def\vv{{\bm{v}}}

\def\vx{{\bm{x}}}
\def\vy{{\bm{y}}}
\def\vz{{\bm{z}}}

\def\mA{{\bm{A}}}

\def\mW{{\bm{W}}}

\DeclareMathAlphabet{\mathsfit}{\encodingdefault}{\sfdefault}{m}{sl}
\SetMathAlphabet{\mathsfit}{bold}{\encodingdefault}{\sfdefault}{bx}{n}

\newcommand{\R}{\mathbb{R}}

\newcommand{\softmax}{\mathrm{softmax}}

\theoremstyle{plain}

\theoremstyle{definition}

\theoremstyle{remark}

\usepackage{xurl}

\usepackage[textsize=tiny]{todonotes}

\icmltitlerunning{Gated Linear Attention Transformers with Hardware-Efficient Training}

\begin{document}

\twocolumn[
\icmltitle{Gated Linear Attention Transformers with Hardware-Efficient Training}

\icmlsetsymbol{equal}{*}

\begin{icmlauthorlist}
\icmlauthor{Songlin Yang}{yyy,equal}
\icmlauthor{Bailin Wang}{yyy,equal}
\icmlauthor{Yikang Shen}{comp}
\icmlauthor{Rameswar Panda}{comp}
\icmlauthor{Yoon Kim}{yyy}
\end{icmlauthorlist}

\icmlaffiliation{yyy}{Massachusetts Institute of Technology}
\icmlaffiliation{comp}{MIT-IBM Watson AI Lab}

\icmlcorrespondingauthor{Songlin Yang}{\url{yangsl66@mit.edu}}
\icmlcorrespondingauthor{Bailin Wang}{\url{bailinw@mit.edu}}

\icmlkeywords{Machine Learning, ICML}

\vskip 0.3in
]

\printAffiliationsAndNotice{\icmlEqualContribution} %
\vspace{-2mm}
\begin{abstract}
\vspace{-1mm}
Transformers with linear attention allow for efficient parallel training  but can simultaneously  be formulated as an RNN with 2D (matrix-valued) hidden states, thus enjoying  linear-time inference complexity. However, linear attention generally underperforms ordinary softmax attention. Moreover, current implementations of linear attention  lack I/O-awareness and are thus slower than highly optimized implementations of softmax attention. This work  describes a hardware-efficient algorithm for linear attention that trades off memory movement against parallelizability. The resulting implementation, dubbed \textsc{FlashLinearAttention}, is faster than \textsc{FlashAttention-2} \cite{flashattention2} as a standalone layer even on short sequence lengths (e.g., 1K). We then generalize this algorithm to 
 a more expressive variant of linear attention with data-dependent gates. When used as a replacement for the standard attention layer in Transformers, the resulting gated linear attention (GLA) Transformer is found to perform  competitively against the LLaMA-architecture Transformer \citep{touvron2023llama} as well recent linear-time-inference baselines such as RetNet \citep{sun2023retentive} and Mamba \citep{Gu2023MambaLS} on moderate-scale language modeling experiments.  GLA Transformer is especially effective at length generalization, enabling a model trained on 2K  to generalize to sequences longer than 20K without significant perplexity degradations. For training speed, the GLA Transformer has higher throughput than a similarly-sized Mamba model.

\begin{minipage}{\linewidth}

   \faGithub \quad  \url{https://github.com/sustcsonglin/flash-linear-attention}
\end{minipage}

\end{abstract}

\vspace{-2mm}
\section{Introduction}
\vspace{-4mm}
\label{sec:intro}
Transformers with softmax attention \citep{vaswani2017attention}  enjoy efficient parallel training but suffer from quadratic (in sequence length) complexity, thus motivating more RNN-like models that allow for linear-time sequence modeling. Linear attention, which replaces the exponential similarity function with a simple dot product over (possibly transformed) key/query vectors, has emerged as a promising alternative to classic softmax attention \citep{katharopoulos2020transformers,performer,kasai-etal-2021-finetuning,peng2021random}. An attractive property of linear attention  is that it admits a  ``recurrent form'' in which it can be formulated as a linear RNN with 2D hidden states \citep{katharopoulos2020transformers}, thus enabling linear-time  inference. For training, linear attention  also admits a subquadratic ``chunkwise parallel form'' which divides the sequence into non-overlapping chunks and performs (serial) inter-chunk recurrent computations followed by (parallel) intra-chunk computations \citep{GAU,sun2023retentive,VQ-Transformer}, thus (partially) maintaining  parallel training. However,  existing algorithms for linear attention are not I/O aware and thus, in practice, slower than  optimized implementations of softmax attention \cite{flashattention1,flashattention2} on moderate sequence lengths.

From a performance standpoint,  linear attention has generally been found to underperform ordinary softmax attention, often by a significant margin in language modeling \cite{kasai-etal-2021-finetuning}. Recent variants of linear attention such as RetNet \citep{sun2023retentive} and TransNormerLLM \citep{qin2023scaling} obtain significant improvements by multiplying the current hidden state with a decay factor before the RNN update. However,  these works use a global, \emph{data-independent} decay factor, despite the fact that in 1D RNNs, a \emph{data-dependent} gating mechanism has been shown to be crucial for performance \citep{unreasonable-forget-gate,HGRN}. And even with the decay factor, linear attention Transformers underperform the strongest Transformer architectures when pretrained from scratch. 

This work develops a hardware-efficient algorithm for linear attention, and applies it to train a gated variant of linear attention that is competitive with softmax attention. We first discuss  aspects of optimizing ordinary linear attention on modern GPUs  and give two I/O-aware algorithms (tailored for different training settings) based on these principles (\S\ref{sec:algorithm}). Our implementation of the algorithm, called \textsc{FlashLinearAttention}, is faster than \textsc{FlashAttention-2} \citep{flashattention2} even on short (e.g., 1K) sequences. 
We then describe a gated linear attention layer with a data-dependent gating mechanism and show how \textsc{FlashLinearAttention} can be generalized to the gated case (\S\ref{sec:gla}). 
We study the resulting \emph{gated linear attention (GLA) Transformer} on moderate-scale language modeling benchmarks, where we train models with 340M/1.3B parameters on 15B/100B tokens, respectively. We find that the GLA Transformer performs favorably against a strong LLaMA architecture Transformer baseline that makes use of recent recipes \citep[Transformer++;][]{touvron2023llama} as well as recent linear-time sequence models such as RetNet \cite{sun2023retentive} and Mamba \citep{Gu2023MambaLS}. GLA Transformer is found to be particularly strong at length generalization and recall-intensive tasks among linear recurrent models. For training speed,  the GLA Transformer has significantly  higher throughput than a similarly sized Mamba model.

\vspace{-3mm}
\section{Background: Linear Attention}
\vspace{-3mm}
We first give a brief background on linear attention layers. For notation we  use bold upper-case letters for matrices (e.g., $\rmS$, $\rmQ$),  bold lower-case letters  for vectors (e.g., $\vq_t$, $\vk_t$), and italic upper-case  for learnable parameters matrices (e.g., $\mW_K$). We generally use the same alphabet to show the rows of a matrix, e.g., $\vq_t$ is the $t$-th row of $\rmQ$.

\label{sec:background}
\vspace{-2mm}
\subsection{Parallel and Recurrent Forms}
\vspace{-2mm}
\label{subsec:background-lin}
Standard autoregressive Transformers employ a softmax attention mechanism which takes an input sequence $\rmX \in \R^{L \times d}$ (here $L$ is the length and $d$ is the hidden dimension) and computes the output $\rmO \in \R^{L \times d}$ through,
\begin{align*}
\rmQ, \rmK, \rmV &= \rmX \mW_Q, \rmX \mW_K, \rmX \mW_V, \\  \rmO &= \softmax\big((\rmQ \rmK^\intercal) \odot \rmM \big) \rmV,
\end{align*}
where $\mW_Q, \mW_K, \mW_V \in \R^{d \times d}$ are learnable matrices and 
$\rmM \in \{-\infty,1\}^{L \times L}$ is a mask that prevents the model from attending to  future tokens, i.e., $\rmM_{ij}=1$ if $i\ge j$ and $\rmM_{ij}=-\infty$ if $i<j$. (Here we assume a single attention head for simplicity.) 
The above \textit{parallel form} of attention can compute $\rmO$ in parallel given the full input $\rmX$, thus enabling efficient training. However, during inference Transformers must use the following \emph{recurrent form},
\begin{align*}
\vq_t, \ \vk_t, \ \vv_t &= \vx_t \mW_Q, \  \vx_t \mW_K, \ \vx_t \mW_V, \\ \vo_t &= \frac{\sum_{i=1}^{t} \exp(\vq_t  \vk_i^\intercal)\vv_i}{\sum_{i =1} ^{t} \exp(\vq_t  \vk_i^\intercal)},
\end{align*}
which calculates the query ($\vq_t$), key ($\vk_t$), and value ($\vv_t$) vectors given the current token's representation $\vx_t \in \R^{1 \times d}$ and the performs attention over the (growing) set of keys $\{\vk_1, \dots, \vk_t\}$ and values   $\{\vv_1, \dots, \vv_t\}$ (i.e., the ``KV cache''). 

Linear attention mechanisms~\citep{katharopoulos2020transformers} replace $\exp(\vq_t \vk_i^\intercal)$ with a kernel $k(\vx, \vy)$ with an associated feature map $\phi$ (i.e., $k(\vx, \vy) = \langle\phi(\vx), \phi(\vy)\rangle$).  This simplifies the calculation of $\vo_t$ since we have
\begin{align*}
 \vo_t &= \frac{\sum_{i=1}^{ t}\phi(\vq_t)\phi(\vk_i)^\intercal \vv_i}{\sum_{i=1}^{t} \phi(\vq_t)\phi(\vk_i)^\intercal  }  
 = \frac{\phi(\vq_t)   \sum_{i=1}^{t}\phi(\vk_i)^\intercal \vv_i}{\phi(\vq_t) \sum_{i=1}^{t}\phi(\vk_i)^\intercal}.
\end{align*}
Letting $\rmS_t=\sum_{i=1}^{t}\phi(\vk_i)^\intercal \vv_i$ and $\vz_t=\sum_{i=1}^{t}\phi(\vk_i)^\intercal$ where $\rmS_t \in \mathbb{R}^{d\times d}, \vz_t \in \mathbb{R}^{d\times 1}$, we can rewrite the above as an RNN,
\begin{align*}
\rmS_t = \rmS_{t-1} &+ \phi(\vk_t)^\intercal \vv_t, \hspace{1mm} \vz_t = \vz_{t-1} + \phi(\vk_t)^\intercal, \hspace{1mm} \vo_t = \frac{\phi(\vq_t) \rmS_t}{ \phi(\vq_t) \vz_t}.
\end{align*}
 Although various kernels have been explored~\citep{kasai-etal-2021-finetuning,peng2021random}, recent work has found that a linear kernel (i.e., setting $\phi$ to be the identity) without a normalizer  works well in practice \cite{sun2023retentive}. 
This results in an (unnormalized) linear attention layer with the following update equation,
\begin{align}
& \rmS_t = \rmS_{t-1} + \vk_t^\intercal \vv_t, \quad \vo_t = \vq_t \rmS_t   .
\label{eq:simple_linear_attention}
\end{align}
Eq.~\ref{eq:simple_linear_attention} makes it clear that  a linear attention layer is essentially a linear recurrent layer with matrix-valued hidden states $\rmS_t$ that is updated via the outer-product $\vk_t^\intercal\vv_t = (\vx_t \mW_K)^\intercal (\vx_t \mW_V)$.\footnote{This type of model with matrix-valued hidden states that change over time is also known as ``fast weights'' \citep{hinton1987using,schmidhuber1992learning,ba2016using}, whose connection to Transformers was  explored in recent work \citep{linear-xmr-fastweight,irie2021going,mao-2022-fine}.} 
The parallel form of causal linear attention, whose complexity is still quadratic in $L$, is given by
$\rmO = \big((\rmQ \rmK^\intercal) \odot \rmM \big)\rmV$,
where $\rmM \in \{0,1\}^{L \times L}$ is a mask such that $\rmM_{ij}=1$ if $i\ge j$ and $\rmM_{ij}=0$ if $i<j$. Due to $\rmM$ it is not possible to exploit the associative property of matrix multiplication to reduce the parallel form complexity from quadratic to linear.\footnote{Without $\rmM$, one can transform $(\rmQ \rmK^\intercal)\rmV$ to $\rmQ( \rmK^\intercal\rmV)$ reducing the complexity from quadratic ($O(L^2d)$) to linear ($O(Ld^2)$).}

\vspace{-3mm}
\subsection{Chunkwise Parallel Form} 
\vspace{-2mm}
\label{background:lin-chunkwise}

The \emph{chunkwise} parallel form of linear attention  strikes a balance between parallel and recurrent form \cite{GAU,sun2023retentive}, and allows for subquadratic, partially parallel training.
 Formally, suppose the input $\rmX$ is now split into non-overlapping chunks, where each chunk is of length $C$. 
Let $\rmS_{[i]} \in \R^{d \times d}$ be the chunk-level hidden state after processing $i$ chunks, i.e., $\rmS_{[i]}:=\rmS_{iC}$. Further let $\rmQ_{[i]}:=\rmQ_{iC+1:(i+1)C+1} \in \mathbb{R}^{C\times d}$ be the query vectors corresponding to the $i$-th chunk; let $\rmK_{[i]}$, $\rmV_{[i]}$, $\rmO_{[i]}$ be similarly defined. We then have the following inter-chunk recurrence (for $i \in [0, 1, \dots \frac{L}{C}-1]$):
    \begin{equation}
\rmS_{[i+1]} = \rmS_{[i]} + \underbrace{\sum_{j=iC + 1}^{(i+1)C} \vk_{j}^\intercal \vv_{j}}_{\rmK^\intercal_{[i]}\rmV_{[i]}} \quad \hspace{1mm} \in \mathbb{R}^{d\times d}.
\label{eq:la-inter-chunk}
    \end{equation}
Here  $\rmS_{[0]}$ can be initialized to  zero or from the previous segment's hidden state. The sum of all RNN inputs from a chunk (i.e., $\rmK^\intercal_{[i]}\rmV_{[i]}$) can be computed in $O(C^2d)$ in parallel.
The intra-chunk parallel computation for the output is given by
\begin{equation*}    
\rmO_{[i+1]} = \underbrace{\rmQ_{[i+1]}\rmS_{[i]}}_{\text{inter-chunk}: \rmO^\text{inter}_{[i+1]}} + \underbrace{\big((\rmQ_{[i+1]}\rmK_{[i+1]}^{\intercal})\odot\rmM\big)\rmV_{[i+1]}}_{\text{intra-chunk}: \rmO^{\text{intra}}_{[i+1]}},
\label{eq:la_block_wise}
\end{equation*}
where $\rmO_{[i+1]} \in \mathbb{R}^{C\times d}$. Here the ``intra-chunk'' component $\rmO^\text{intra}_{[i+1]}$  has exactly the same parallel form as Eq.~\ref{eq:simple_linear_attention} and thus takes $O(C^2d + Cd^2)$. The ``inter-chunk'' component $\rmO^\text{inter}_{[i+1]}$ accounts for the contribution from the hidden state from the previous chunk, and takes $O(Cd^2)$. Training complexity is thus  $O\left(\frac{L}{C}(C^2d + Cd^2) \right)=O(LCd+Ld^2)$, which is less than $O(L^2d)$ when $L>d$.  Note that setting $C = L$ recovers the parallel form, and $C=1$ recovers the recurrent form.

\vspace{-2mm}
\section{Hardware-Efficient Linear Attention}
\vspace{-2mm}
\label{sec:algorithm}
We describe \textsc{FlashLinearAttention}, an I/O-aware, hardware-efficient algorithm for linear attention in the spirit of \textsc{FlashAttention} \cite{flashattention1,flashattention2}. We first discuss aspects of hardware that should be taken into account for a practically efficient implementation.

\vspace{-2mm}
\subsection{Principles of Hardware-Efficient Algorithms}
\vspace{-2mm}
 An efficient  algorithm should be aware of the compute model, memory hierarchy, and specialized compute units on modern hardware.

\vspace{-2mm}
\paragraph{Occupancy.} GPUs have many threads executed in parallel; threads are grouped into thread blocks, which execute on streaming multiprocessors (SMs). To maintain a high GPU occupancy (i.e., fraction of GPU resources being used), it is necessary to use a sufficient number of SMs. 
In large-scale training and long-sequence modeling scenarios where the batch size tends to be small, 
parallelizing over the temporal dimension enables high GPU occupancy \cite{flashattention2}.

\vspace{-2mm}
\paragraph{Specialized compute units.} Modern hardware for neural network training typically have specialized compute units (e.g., {tensor cores} on NVIDIA GPUs, matrix mutiply units on TPUs), which can significantly accelerate matmuls; for example half-precision matmuls
on an A100 can be roughly 16 times faster on tensor cores than on CUDA cores. These specialized units are crucial for large-scale training.

\vspace{-2mm}
\paragraph{Memory hierarchy.} GPUs have a memory hierarchy with larger but slower global GPU memory (high bandwidth memory; HBM) and smaller but faster shared memory (SRAM). Optimal utilization of SRAM to reduce HBM I/O cost can therefore lead to significant speed-ups.

\begin{algorithm}[t!]
\vspace{-0.5mm}
\scriptsize
\caption{\textsc{FlashLinearAttention}: Forward Pass}
\label{algo:la-chunk-fwd}
\begin{algorithmic}
    \Require $\rmQ, \rmK, \rmV \in \R^{L \times d}, \rmV  \in \R^{L \times d}$, chunk size $C \in [L]$, \texttt{materialize} $\in$ \{\texttt{True,False}\} 
    \State Divide $\rmQ, \rmK, \rmV$ into $N = \frac{L}{C}$ blocks $\{ \rmQ_{[1]} \dots \rmQ_{[N]} \} $, $\{ \rmK_{[1]} \dots \rmK_{[N]} \}$ of size $C \times d$ each.
     \State{Initialize $\rmS = \bm{0} \in \mathbb R^{d\times d} $ on SRAM}
    \State{On chip, construct causal mask $\rmM\in \R^{C\times C}$}
    \If{\texttt{materialize}} \Comment{the materialization version}
    \For{$n \gets 1, N$}
      \State{Store $\rmS$ to HBM as 
      $\rmS_{[n]}$.}
        \State{Load $\rmK_{[n]}, \rmV_{[n]} \in \mathbb{R}^{C \times d}$ from HBM to SRAM}
        \State{On chip, compute $\mathbf{S} = \mathbf{S} + \rmK_{[n]}^{\top}  \rmV_{[n]}$.}
        \color{black}
    \EndFor
    \ParFor{$n \gets 1, N$}
        \State Load $\rmQ_{[n]}, \rmK_{[n]}, \rmV_{[n]}, \rmS_{[n]}$ from HBM to SRAM.
        \State On chip, compute $\rmO^{\prime} = \rmQ_{[n]} \rmS_{[n]} + (\rmQ_{[n]} \rmK^\intercal_{[n]} \odot \rmM)\rmV_{[n]}$  
        \State Store $\rmO^{\prime}$ to HBM as $\rmO_{[n]}$.
    \EndParFor
  \State \Return $\rmO = \{ \rmO_{[1]} \dots \rmO_{[N]} \}$, $\rmS=\{ \rmS_{[1]} \dots \rmS_{[N]} \}$.
        \Else \Comment{the non-materialization version}
         \For{$n \gets 1, N$}
      \color{black}
        \State{Load $\rmQ_{[n]}, \rmK_{[n]}, \rmV_{[n]} \in \mathbb{R}^{C \times d}$ from HBM to SRAM}
        \State{On chip, compute $\rmO' = \rmQ_{[n]}\rmS + (\rmQ_{[n]}\rmK_{[n]}^{\top}\odot\rmM)\rmV_{[n]}$}
        \State{On chip, compute $\mathbf{S} = \mathbf{S} + \rmK_{[n]}^{\top}  \rmV_{[n]}$.}
        \State{Store $\rmO'$ to HBM as $\rmO_{[n]}$ }
        \color{black}
    \EndFor
   \State  \Return $\rmO = \{ \rmO_{[1]} \dots \rmO_{[N]} \}$   
\EndIf 
\vspace{-0.5mm}
\end{algorithmic}
\end{algorithm}

\vspace{-2mm}
\subsection{Hardware Considerations for Linear Attention}
\vspace{-2mm}
We now discuss hardware considerations pertaining to the efficiency of the different forms of linear attention.  

\vspace{-2mm}
\paragraph{Recurrent form.}  A basic implementation of the recurrent form  stores the 2D hidden states of all time steps in HBM, resulting in  high I/O cost \cite{mao-2022-fine}. I/O cost could be reduced by avoiding such materialization and recomputing the hidden states during the backward pass, as in \citet{katharopoulos2020transformers}, but the elementwise operations in the recurrent update cannot make use of tensor cores and result in low arithmetic intensity. Hence, while the recurrent form generally has the lowest total FLOPs among the three forms, this does not translate to actual wall-time efficiency.  And while it is theoretically possible to parallelize  linear recurrences via the parallel scan algorithm, this method requires materializing the 2D hidden state for each time step. This  incurs a significant memory I/O burden, thereby offsetting the benefits of parallelism over the sequence length and resulting in slow actual running speeds, as in 
 \citet{gatedloop}.

\vspace{-2mm}
\paragraph{Parallel form.} The parallel form could be as efficient as \textsc{FlashAttention} using similar I/O optimization techniques, as demonstrated by \citet{qin2023scaling}. However, the high number of FLOPs (due to the quadratic complexity) makes the long-sequence training expensive, the same issue that the na\"ive implementation of softmax attention would suffer from. %

\vspace{-2mm}
\paragraph{Chunkwise form.} The chunkwise parallel form, which interpolates between the parallel and recurrent forms with an extra ``parameter'' $C$, makes it possible to more easily make the above tradeoffs for fine-grained  optimization. 
Unlike the recurrent form, most operations can be done via matmuls, enabling the use of tensor cores (if $C$ is set to a multiple of 16). 
Though the chunkwise training algorithm has been discussed before in the literature \cite{GAU, sun2023retentive}, most implementations are not I/O-aware and thus slower than \textsc{FlashAttention} for moderate sequence lengths (e.g., 2K-4K). 

\usetikzlibrary{arrows.meta,
                positioning,
                shadows}

\usetikzlibrary{shapes.multipart,positioning}

\usetikzlibrary{shapes, arrows, arrows.meta, fit,backgrounds, positioning, calc}

\tikzstyle{process} = [rectangle, minimum width=2cm, minimum height=1cm, text centered, text width=2cm,  draw=black, fill=orange!30]
\tikzstyle{arrow} = [thick,->,>=stealth]
\tikzstyle{container1} = [draw, rectangle, inner sep=0.3cm, fill=gray!20]
\tikzset{
  mybackground/.style={execute at end picture={
      \begin{scope}[on background layer]
        \node[] at (current bounding box.north){\bottom{1cm} #1};
        \end{scope}
    }},
}

\tikzset{
    font=\sffamily,
    BLOCK/.style={
        draw,
        align=center,
        draw=red!50,
        fill=red!20,
        rectangle split, 
        rectangle split horizontal,
        rectangle split parts=#1, 
    }
}
\tikzset{cross/.style={cross out, draw, 
         minimum size=2*(#1-\pgflinewidth), 
         inner sep=0pt, outer sep=0pt}}
\begin{figure}[t]
\centering  
		\resizebox{.8\columnwidth}{!}{  
        \begin{tikzpicture}[%
            every path/.style={thick,%
            },
                           dcs/.style = {double copy shadow, shadow xshift=4pt, shadow yshift=-4pt}
          ]
        \node [rectangle, minimum width=.5cm, minimum height=1cm, text centered, text width=1cm,  draw=black, fill=blue!30] (process3) at (-2, -1.5) {$\mathbf{S}_{[i-1]}$};

        \node [rectangle, minimum width=.5cm, minimum height=.75cm, text centered, text width=.75cm,  draw=black, fill=red!20] (hh1) at (-2, -6.5) {$\mathbf{S}_{[i-1]}$};
        \node [rectangle, minimum width=.5cm, minimum height=.75cm, text centered, text width=.75cm,  draw=black, fill=red!20] (hh2) at (-0.5, -6.5) {$\mathbf{S}_{[i]}$};
        \node [rectangle, minimum width=.5cm, minimum height=.75cm, text centered, text width=.75cm,  draw=black, fill=red!20] (hh3) at (1, -6.5) {$\mathbf{S}_{[i+1]}$};

        \node [rectangle, minimum width=.5cm, minimum height=.7cm, text centered, text width=.7cm,  draw=black, fill=orange!30] (hh4) at (3, -6.5) {$\mathbf{S}_{[i]}$};
        
        \node [rectangle, minimum width=.5cm, minimum height=1cm, text centered, text width=1cm,  draw=black, fill=blue!30] (h2) at (2, -1.5) {$\rmS_{[i]}$};
        \node [rectangle, minimum width=.5cm, minimum height=1cm, text centered, text width=1cm,  draw=black, fill=blue!30] (h3) at (6, -1.5) {$\rmS_{[i+1]}$};
        
        \node[draw, align=center,        draw=red!50,fill=red!20,rectangle split, 
        rectangle split horizontal,
        rectangle split parts=2, 
    ] (block4) at (4, 0){
        \nodepart{one} $\rmO^{\text{inter}}_{[i+1]}$ \nodepart{two} $\rmO^{\text{intra}}_{[i+1]}$ };
            
\node[draw,
        align=center,
        draw=red!50,
        fill=red!20,
        rectangle split, 
        rectangle split horizontal,
        rectangle split parts=2, 
    ] (blockk4) at (5.5, -6.5){
        \nodepart{one} $\rmO^{\text{inter}}_{[i+1]}$ \nodepart{two} $\rmO^{\text{intra}}_{[i+1]}$
    };

    \node[draw,
        align=center,
        draw=red!50,
        fill=red!20,
        rectangle split, 
        rectangle split horizontal,
        rectangle split parts=2, 
    ] (block2) at (0,0)
    {
        \nodepart{one} $\rmO^{\text{inter}}_{[i]}$ \nodepart{two} $\rmO^{\text{intra}}_{[i]}$
    };
    
    \node[BLOCK=3, fill=orange!30, draw=orange!80
    ] (block) at (0,-3) {
        \nodepart{one} $\rmQ_{[i]}$ \nodepart{two} $\rmK_{[i]}$ \nodepart{three} $\rmV_{[i]}$
};

    \node[BLOCK=2, fill=orange!30, draw=orange!80
    ] (blockk1) at (-2,-8) {
      \nodepart{one} $\rmK_{[i]}$ \nodepart{two} $\rmV_{[i]}$
};
    \node[BLOCK=2, fill=orange!30, draw=orange!80
    ] (blockk2) at (0.5,-8) {
      \nodepart{one} $\rmK_{[i+1]}$ \nodepart{two} $\rmV_{[i+1]}$
};
    \node[BLOCK=3,  fill=orange!30, draw=orange!80
    ] (block3) at (4,-3) {
        \nodepart{one} $\rmQ_{[i+1]}$ \nodepart{two} $\rmK_{[i+1]}$ \nodepart{three} $\rmV_{[i+1]}$
        
    };
    \node[BLOCK=3,  fill=orange!30, draw=orange!80
    ] (blockk3) at (4.75,-8) {
        \nodepart{one} $\rmQ_{[i+1]}$ \nodepart{two} $\rmK_{[i+1]}$ \nodepart{three} $\rmV_{[i+1]}$
    };

    \node [rectangle, text centered, text width=.5cm, text height=.5cm, draw=black, fill=orange!30, label=right:{Load from HBM}] (node1) at (-2.5, -4.25){};
    \node [rectangle, text centered, text width=.5cm, text height=.5cm, draw=black, fill=red!20, label=right:Store to HBM] (node2) at (.9, -4.25){};
    \node [label=right:(a)] (labell) at (6, -3.3){};
        \node [label=right:(c)] (labelc) at (6, -8.6){};
            \node [label=right:(b)] (labelb) at (1, -8.6){};
    \node [rectangle, text centered, text width=.5cm, text height=.5cm, draw=black, fill=blue!30, label=right:On-chip construct] (node3) at (4, -4.25){};

            \draw [arrow] (process3) -- (block2.one south);
                \draw [arrow] (h2) -- (block4.one south);
            \draw [arrow] (block.one north) -- (block2.one south);
            \draw [arrow, dashed] (process3) -- (h2);
            \draw [arrow, dashed] (h2) -- (h3);
            \draw [arrow] (block.one north) -- (block2.two south);
                    \draw [arrow] (block3.one north) -- (block4.one south);
                                    \draw [arrow] (block3.two north) -- (block4.two south);
            \draw [arrow] (block3.three north) -- (block4.two south);
            \draw [arrow] (block3.three north) -- (h3);
            \draw [arrow] (block3.one north) -- (block4.two south);
            \draw [arrow] (block.two north) -- (block2.two south);
            \draw [arrow] (block.three north) -- (block2.two south);
            \draw [arrow] (block.two north) -- (h2);
            \draw [arrow] (block.three north) -- (h2);
            \draw [arrow, dashed] (hh1) -- (hh2);
            \draw [arrow, dashed] (hh2) -- (hh3);
            \draw [arrow] (blockk1.one north) -- (hh2);
            \draw [arrow] (blockk1.two north) -- (hh2);
            \draw [arrow] (blockk2.two north) -- (hh3);
            \draw [arrow] (blockk2.one north) -- (hh3);
            \draw [arrow] (hh4) -- (blockk4.one west);
            \draw [arrow] (blockk3.one north) -- (blockk4.one west);
                        \draw [arrow] (blockk3.one north) -- (blockk4.two south);
            \draw [arrow] (blockk3.two north) -- (blockk4.two south);
            \draw [arrow] (blockk3.three north) -- (blockk4.two south);
                        \draw [arrow, dashed] ($(hh1)-(1,0)$) -- (hh1);
                        \draw [arrow, dashed] (hh3) -- ($(hh3)+(1,0)$);
                        \draw [arrow, dashed] (h3) -- ($(h3)+(1.5,0)$);
                        \draw [arrow, dashed] ($(process3)-(1.5,0)$) -- (process3);

        \begin{scope}[on background layer]
    \node  [container1,fit= {(process3) (h2) (block) (block2) (h3) (block3) (labell)}, label=north:\large Sequential] (contain1) {};
    \end{scope}
                        \begin{scope}[on background layer]
                \node  [container1,fit= { (hh1) (hh2) (hh3) (blockk1)  (blockk2) (labelb)}, label=north:\large Sequential] (contain2) {};
            \end{scope}
                        \begin{scope}[on background layer]
                \node  [container1,dcs, fit= { (hh4) (blockk3) (blockk4) (labelc)}, label={[label distance=0.25cm]north:\large Chunkwise Parallel}, fill=yellow!20] (contain3) {};
            \end{scope}

        \end{tikzpicture}   
}
        \vspace{-2mm}
        \caption{(a) \textsc{FlashLinearAttention} without materialization. This version is more memory-efficient. (b-c) \textsc{FlashLinearAttention} with materialization. This version enables sequence-level chunkwise parallelism.} \label{fig:chunk}
              
    \end{figure}
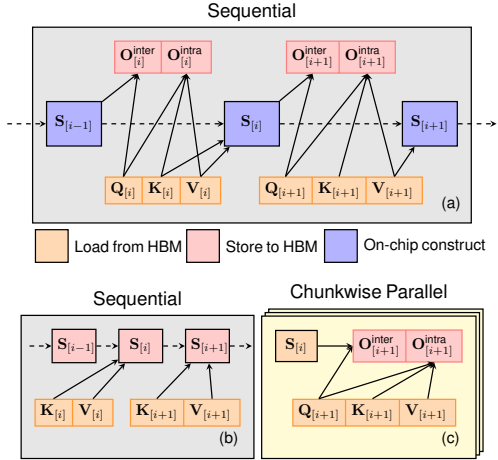

\vspace{-2mm}
\subsection{\textnormal{\textsc{FlashLinearAttention}}: Hardware-Efficient  Linear Attention with the Chunkwise Form}
\label{sec:fla}
\vspace{-2mm}
We describe our I/O-aware, hardware-efficient implementation of the chunkwise form. We give two versions, whose forward and backward passes differ depending on whether the chunk-level hidden states  $\rmS_{[n]}$ are materialized in HBM. See Alg.~\ref{algo:la-chunk-fwd} and Fig.~\ref{fig:chunk} for the forward pass. (Alg.~\ref{algo:la-chunk-bwd} in the appendix describes the backward pass.)  At a high level, we use tiling to load tensors block-by-block and re-use tensor blocks on chip to avoid multiple HBM I/O as {much} as possible. For example, when $\rmQ_{[n]}$ is loaded to SRAM, both $\rmQ_{[n]}\rmS$ and $(\rmQ_{[n]}\rmK_{[n]}^{\top}\odot\rmM)\rmV_{[n]}$ can be computed on chip, which avoids loading $\rmQ_{[n]}$ twice, thus saving HBM I/O.

\textbf{The non-materialization version} computes $\rmO_{[n]}$ sequentially for $n \in [N]$, using SRAM to temporarily store $\rmS_{[n]}$, which is memory-efficient. This version parallelizes across batch size, number of heads, and head dimensions, but lacks sequence-level parallelim. When the batch size is large, this level of parallelism is sufficient to enable high GPU occupancy. In long-sequence and large scale training settings where batch size is small, the SMs cannot be fully exploited in this case. \textbf{The materialization version} first performs the inter-chunk recurrence (Eq.~\ref{eq:la-inter-chunk}) and stores all $\rmS_{[n]}$ for $n \in [N]$ in HBM. Then, the $\rmO_{[n]}$'s can be computed in parallel for all chunks.  
 This approach offers better parallelism but increases the memory footprint by approximately 10-20\%. We mitigate this through \textit{recomputation}, where the hidden states discarded  after the forward pass and recomputed during the backward pass. We find this introduces a small runtime  overhead but significantly reduces the memory footprint, and  we adopt this strategy by default.

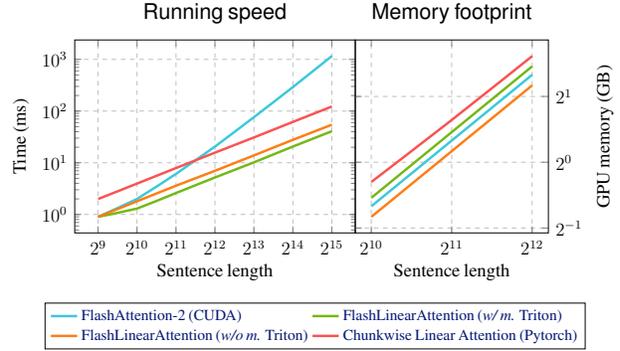
\begin{figure}[t]
\centering  
		\resizebox{\columnwidth}{!}{  
\begin{tikzpicture}
\scalefont{1.4} %
\begin{axis}[%
name=plot1,
title={Running speed},
scatter/classes={%
1={mark=diamond*,draw=black},
0={mark=o,draw=black}},
legend style={at={(5.5,-1.6)},   
                anchor=north,legend columns=2,
                column sep=0cm,
                font=\normalsize,
                row sep=0cm
                },     
    legend cell align=left,
    tick label style={font=\large},
    label style={font=\large},
legend to name={mylegend},
			ymajorgrids=true, %
			xmajorgrids=true, %
			grid style=dashed, %
xmode=log,
ymode=log,
log basis x={2},
ylabel={Time (ms)},
xlabel={Sentence length},
width=8cm,
height=6cm,
every axis plot/.append style={ultra thick}
]
\addplot[smooth,bcyan,tension={0.15}]
coordinates{
            (512,0.9)
            (1024,2.0)
            (2048,6.1)
            (4096,20.6)
            (8192,76.3)
            (16384,291.4)
            (32768,1160.3)
            };
\addplot[smooth,bgreen,tension={0.15}]
coordinates{
            (512,0.9)
            (1024,1.3)
            (2048,2.6)
            (4096,5.2)
            (8192,10.2)
            (16384,20.6)
            (32768, 40.8)
            };
\addplot[smooth,borange,tension={0.15}]
coordinates{
            (512,0.9)
            (1024,1.8)
            (2048,3.6)
            (4096,7.0)
            (8192,13.9)
            (16384,27.8)
            (32768, 54.5)
            };
\addplot[smooth,bred,tension={0.15}]
coordinates{
            (512,1.99)
            (1024,3.96)
            (2048,7.91)
            (4096,15.71)
            (8192,30.92)
            (16384,61.79)
            (32768, 122.55)
            };
\legend{FlashAttention-2 (CUDA), FlashLinearAttention (\textit{w/ m.} Triton),  FlashLinearAttention (\textit{w/o m.} Triton), Chunkwise Linear Attention (Pytorch) }
\end{axis}  
\begin{axis}[%
at={(plot1.south east) },
title={Memory footprint},
scatter/classes={%
1={mark=diamond*,draw=black},
0={mark=o,draw=black}},
xmode=log,
ymode=log,
    ylabel near ticks, yticklabel pos=right,
log basis x={2},
log basis y={2},
    tick label style={font=\large},
    label style={font=\large},
ylabel={GPU memory (GB)},
xlabel={Sentence length},
			ymajorgrids=true, %
			xmajorgrids=true, %
			grid style=dashed, %
width=6cm,
height=6cm,
    every axis plot/.append style={ultra thick}
]
\addplot[smooth,bcyan,tension={0.15}]
coordinates{
(1024, 0.6289067268371582)
(2048, 1.2578129768371582)
(4096, 2.515625476837158)
};
\addplot[smooth,bgreen,tension={0.15}]
coordinates{
(1024, 0.6875)
(2048, 1.375)
(4096, 2.75)
};
\addplot[smooth,borange,tension={0.15}]
coordinates{
(1024, 0.5625)
(2048, 1.125)
(4096, 2.25)
};
\addplot[smooth,bred]
coordinates{
(1024, 0.8125)
(2048, 1.5625)
(4096, 3.0625)
};
\end{axis}  
\ref{mylegend}
\end{tikzpicture}   
}
\vspace{-6mm}
\caption{Speed comparison on a single H100 GPU with batch size 32, number of heads 16, head dimension 64, and chunk size 64. Both x- and y-axes are on log scale. \textit{w/ m.} and \textit{w/o m.} denotes using \textsc{FlashLinearAttention} \textit{with} or \textit{without materialization} of hidden states in HBM.}\label{fig:flashlinearattn}
\end{figure}

 Figure \ref{fig:flashlinearattn} shows the speed and memory footprint of our implementation. Both versions of \textsc{FlashLinearAttention} are substantially faster than \textsc{FlashAttention-2} \cite{flashattention2} and a pure PyTorch (i.e., I/O-unaware) implementation of chunkwise linear attention, showing the benefits of I/O-awareness. 

\vspace{-2mm}
\section{Gated Linear Attention}
\vspace{-2mm}
\label{sec:gla}

The linear recurrence in Eq.~\ref{eq:simple_linear_attention} does not have a decay term or a forget gate, which has been shown to be crucial in RNNs~\citep{HochSchm97,cho2014learning,unreasonable-forget-gate}. The lack of a decay term makes it difficult for a model to ``forget'' information, and has been hypothesized to be partially responsible for the instability of linear attention in long-context tasks~\citep{buckman2024}. Recent works~\citep{sun2023retentive,qin2023scaling} obtain better performance through incorporating a global, \emph{non-data-dependent}  decay factor\footnote{This can be viewed as linear attention with  ALiBi position encodings \cite{alibi2021}.
In practice these works also incorporate  rotary position embeddings \citep[RoPE;][]{rope}.}
$\gamma \in (0, 1)$ into linear attention:
$ \rmS_t = \gamma \rmS_{t-1} + \vk_t^\intercal \vv_t$.
The use of a single $\gamma$ is designed to preserve the attention-style parallel form for efficient training. 
In this work, we consider a data-dependent gating mechanism for linear attention. We show that despite having a more expressive gating factor, the resulting gated linear attention (GLA) layer still admits a hardware-efficient chunkwise form for efficient training.

\vspace{-2mm}
\subsection{Recurrent and Parallel Form of GLA}
\vspace{-2mm}
\label{ssec:gla}
\begin{table*}[t!]
\footnotesize
\centering
\scalebox{0.8}{
\begin{tabular}{l l l }
\toprule
   Model  &  Parameterization  & Learnable parameters   \\
   \midrule
    Mamba \cite{Gu2023MambaLS} & $\rmG_t = \exp(-( \mathbf{1}^{\intercal}\balpha_t)\odot \exp(\mA)),\quad \balpha_t =\operatorname{softplus}(\vx_t\mW_{\alpha_1}\mW_{\alpha_2})$ & $\mA \in \mathbb{R}^{d_k \times d_v}, \quad \mW_{\alpha_1} \in \mathbb{R}^{d \times \frac{d}{16}}, \quad \mW_{\alpha_2} \in  \mathbb{R}^{\frac{d}{16} \times d_v}$ \\
    Mamba-2 \cite{mamba2} & $\rmG_t = \gamma_t \mathbf{1}^\intercal\mathbf{1}, \quad \gamma_t = \exp\left(-\operatorname{softplus}\left(\vx_t \mW_{\gamma}\right)\exp\left(a\right)\right)$  &  $\mW_{\gamma}
    \in \mathbb{R}^{d \times 1}, \quad a\in \mathbb{R}$ \\
    mLSTM \cite{beck2024xlstm, peng2021random} &  $\rmG_t = \gamma_t \mathbf{1}^\intercal\mathbf{1}, \quad \gamma_t = \sigma\left(\vx_t \mW_{\gamma}\right)$ &  $\mW_{\gamma} \in \mathbb{R}^{d \times 1}$ \\
    Gated Retention \cite{Sun2024YouOC} &
  $\rmG_t = \gamma_t \mathbf{1}^\intercal\mathbf{1}, \quad \gamma_t = \sigma\left(\vx_t \mW_{\gamma}\right)^{\frac{1}{\tau}}$ &  $\mW_{\gamma} \in \mathbb{R}^{d \times 1}$ \\
        DFW \cite{mao-2022-fine,recurrent_linear_xfmr}
        & $\rmG_t = \balpha_t^\intercal \bbeta_t, \quad \balpha_t = \sigma(\vx_t \mW_{\alpha}), \quad \bbeta_t = \sigma(\vx_t \mW_{\beta})$ & $ \mW_{\alpha} \in \mathbb{R}^{d \times d_k}, \quad \mW_{\beta} \in \mathbb{R}^{d \times d_v}$   \\
    GateLoop \cite{gatedloop} & $\rmG_t = \balpha_t^\intercal \mathbf{1}, \quad \balpha_t = \sigma\left(\vx_t\mW_{\alpha_1}\right) \exp\left( \vx_t\mW_{\alpha_2} \mathbf{i} \right) $ & $\mW_{\alpha_1} \in \mathbb{R}^{d\times d_k}, \quad \mW_{\alpha_2}\in \mathbb{R}^{d \times d_k}$ \\
  HGRN-2 \cite{qin2024hgrn2} &  $\rmG_t = \balpha_t^\intercal\mathbf{1}, \quad \balpha_t = \boldsymbol{\gamma} + (1-\boldsymbol{\gamma} ) \sigma(\vx_t \mW_{\alpha})$   & $\mW_{\alpha} \in \mathbb{R}^{d \times d_k}, \quad \boldsymbol{\gamma}  \in (0, 1)^{d_k}$
  \\
  RWKV-6 \cite{peng2024eagle} & $\rmG_t =\balpha_t^\intercal\mathbf{1}, \quad \balpha_t = \exp\left(-\exp\left(\vx_t\mW_{\alpha}\right)\right)$ & $\mW_{\alpha} \in \mathbb{R}^{d \times d_k} $ \\
  Gated Linear Attention (GLA) & $\rmG_t = \balpha_t^\intercal \mathbf{1}, \quad \balpha_t = \sigma\left(\vx_t\mW_{\alpha_1}\mW_{\alpha_2}\right)^{\frac{1}{\tau}}$ & $\mW_{\alpha_1} \in \mathbb{R}^{d\times 16}, \quad \mW_{\alpha_2}\in \mathbb{R}^{16 \times d_k}$ \\
  \bottomrule
\end{tabular}
}
\vspace{1mm}
\vspace{-4mm}
\caption{Gated linear attention formulation of recent  models, which vary in their parameterization of $\rmG_t$. The bias terms are omitted.}
\vspace{-4mm}
\label{tab:G_param}
\end{table*}

\paragraph{Recurrent form.} GLA has a 2D forget gate \(\rmG_t \in (0,1)^{d_k \times d_v}\) that varies over time:
\[
\rmS_t = \rmG_t \odot \rmS_{t-1} + \vk_t^{\top} \vv_t,
\]
where we now allow the hidden state to have varying dimensions.
This Hadamard product-based  recurrent form is very general and encompasses many recent RNNs with 2D hidden states, as listed in Table~\ref{tab:G_param}. 

Central to the design of gated linear attention  is the parameterization of \(\rmG_t\) which requires a balance between \textit{parameter-efficiency}, \textit{state size}, 
and \textit{training efficiency}.
A na\"{i}ve mapping $\vx_t \mapsto \rmG_{t}$ to obtain a data-dependent gating matrix would require a matrix of size $d \cdot d_k \cdot d_v$, which would be parameter-inefficient. \citet{mao-2022-fine} propose a more efficient outer-product-based low-rank parameterization (\(\rmG_t = \balpha_t^{\top} \bbeta_t\)), which requires $d \cdot d_v + d  \cdot d_k$ parameters.\footnote{However, \citet{mao-2022-fine} works with only the recurrent form and materializes the hidden states for all time steps in HBM. In Appendix \ref{sec:ggla} we give a new algorithm that reformulates the model in a matrix-multiply-based parallel form, which can make use of (an extension of) \textsc{FlashLinearAttention} for efficient training.}

In Mamba~\cite{Gu2023MambaLS}, \(\rmG_t\) is obtained by combining a  \emph{data-independent} learnable matrix \(\mA\) with a data-dependent vector $\balpha_t$, which allows the matrix to be full rank. However, this prevents the use of tensor cores because it cannot be reformulated into a matrix-multiply format, as discussed in  \citet{mamba2}. The lack of a compact matrix-multiply form  necessitates the materialization of each time step's hidden states. To reduce high I/O costs, \citet{Gu2023MambaLS} develop a hardware-aware algorithm that materializes the hidden states {exclusively} in  SRAM rather than in HBM. Due to limited SRAM capacity, this approach cannot scale to larger hidden states, which, as we will show in our experiments, results in suboptimal performance on recall-intensive tasks. Mamba-2 \cite{mamba2} addresses this limitation with a more {restricted} gating mechanism: \(\rmG_t = \gamma_t \mathbf{1}^T \mathbf{1}\), where \(\gamma_t \in (0,1)\) is a scalar, which makes it possible to to reformulate the recurrence in matrix-multiply form, enabling the use of tensor cores and larger  state sizes. This \emph{scalar} data-dependent gating is also used in \citet{peng2021random}, \citet{Sun2024YouOC}, and \citet{beck2024xlstm}.

 This paper adopts a middle ground between the scalar and the fully low-rank parameterization by using $\rmG_{t} = \balpha_t ^\top \mathbf{1}$.\footnote{Our preliminary experiments with the $\rmG_t = \balpha_t^\top \bbeta_t$ parameterization resulted in only marginal improvements over $\rmG_{t} = \balpha_t ^\top \mathbf{1}$.}
This results in the following recurrent form,
\begin{align}
\rmS_t =  (\balpha_t^{\top} \mathbf{1}) \odot \rmS_{t-1} + \vk_t^{\top} \vv_t = \text{Diag}(\balpha_t)  \rmS_{t-1} + \vk_t^{\top} \vv_t, 
\label{eq:gla-simple}
\end{align}
where $\balpha_t$ is parameterized via a low-rank linear layer followed by sigmoid on $\vx_t$ (see \S\ref{sec:gla-full}). Note that the above formulation 
 is general and  encompasses several recent RNNs \cite{gatedloop, qin2024hgrn2, peng2024eagle}. Thus, the hardware-efficient GLA implementation (described next) could be directly used or adapted to other models.

\vspace{-2mm}
\paragraph{Parallel form.} 
We now describe a parallel form GLA for parallelizing across sequence length. Unrolling Eq.~\ref{eq:gla-simple} gives
\begin{equation*}
\rmS_t = \sum_{i=1}^{t} \left(\left(\prod_{j=i+1}^{t}\balpha_{j}^{\top}\mathbf{1}\right) \odot \vk_i^{\top} \vv_i \right)    
\label{eq:gla_parallel}
\end{equation*}
Letting $\vb_t := \prod_{j=1}^{t} \balpha_j$, we can rewrite the above as 
\begin{align*}
\vo_t = \vq_t \rmS_t &= \vq_t \sum_{i=1}^{t} \left( \left(\frac{\vb_t}{\vb_i} \right)^{\top} \mathbf{1} \right) \odot \vk_i^{\top} \vv_i \\ &= \sum_{i=1}^{t}(\vq_t \odot \vb_t) \left(\frac{\vk_i}{\vb_i}\right)^{\top} \vv_i
\end{align*}
where the division is element-wise. Letting $\rmB \in (0,1)^{L \times d}$ be the matrix obtained from stacking $\vb_t$'s,  the parallel form is:
\begin{equation*}
\rmO = \left(\left(\underbrace{(\rmQ \odot \rmB) \left(\frac{\rmK}{\rmB}\right)^{\top}}_{\mathbf{P}} \right) \odot \rmM \right) \rmV.
\label{eq:parallel_gla}
\end{equation*}
However, this form is not numerical stable as $\vb_t$  is the cumulative product of gate values in $\balpha_j \in (0, 1)^{1 \times d}$, and thus can be extremely small when $t$ is large, making $ \frac{\mathbf{K}}{\rmB}$ explode. To handle this, we can compute in log space for $\mathbf{P}$,\footnote{This form resembles extrapolatable position encoding~\citep{xpos} in that the term inside the exponential can be viewed as a  \textit{data-dependent} relative position factor. }
\begin{align}
    \vspace{-6mm}
    \mathbf{P}_{ij} = \sum_{k=1}^{d} \mathbf{Q}_{ik} \mathbf{K}_{jk} \, \exp(\log \rmB_{ik}-\log\rmB_{jk}), \quad i \ge j,
    \label{eq:log-semiring}
    \vspace{-6mm}
\end{align}
where $k$ denotes feature indices. 
However, unlike vanilla linear attention, as Eq. \ref{eq:log-semiring} cannot be represented via a standard matmul, and it cannot make use of half-precision matmuls on tensor cores. We will show in \S\ref{subsec:flash-gla} how a secondary-level chunking mechanism can enable the use of half-precision matmuls for most computations while maintaining numerical stability, as illustrated in Figure~\ref{fig:gla_tiling}.

\def\width{16}
\def\hauteur{16}

\definecolor{orange}{RGB}{250,230,200}
\definecolor{gray}{RGB}{220,220,220}
\definecolor{pink}{RGB}{247,206,205}

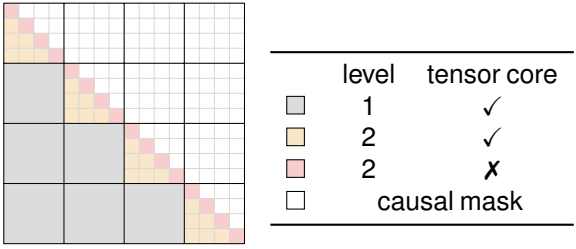
\begin{figure}[t]
\vspace{-2mm}
    \centering
        \begin{tikzpicture}[x=0.2cm, y=0.2cm]

    \foreach \y in {0, ..., 15}{
        \foreach \x in {0, ...,  15} {%
            \pgfmathsetmacro{\col}{ifthenelse(\x<15-\y,"orange",ifthenelse(\x==\y+1,"white","white"))}
       \fill[\col] (\x,\y) rectangle (\x+1,\y+1);
       }
    }
    \foreach \y in {0, ..., 15}{
       \fill[pink] (\y,15-\y) rectangle (\y+1,15-\y+1);
       }

    \draw[step=0.20cm, line width=0.05mm, black!15!white, fill=orange] (0,0) grid (\width,\hauteur);
    \foreach \x in {0, ..., 8} {%
    \fill[gray] (\x,0) rectangle (\x+4,4);
    }
        \foreach \x in {0, ..., 4} {%
    \fill[gray] (\x,4) rectangle (\x+4,8);
    }
    \foreach \x in {0, ..., 1} {%
    \fill[gray] (\x,8) rectangle (\x+3,12);
    }
    
    \foreach \x in {0, ..., 1} {%
    \fill[gray] (\x,8) rectangle (\x+3,12);
    }

    \draw[step=0.8cm, line width=0.1mm, black!10!black, fill=orange] (0,0) grid (\width,\hauteur);

        \node 
        [shape=rectangle,fill=white, 
        align=center](table2) at (28.0,7.0) {

            \begin{tabular}{lcc} \toprule
                 & level & tensor core   \\ 
            \begin{tikz}
                \node [draw,fill=gray] {};
            \end{tikz} & 1 & $\checkmark$\\
                \begin{tikz}
                \node [draw,fill=orange] {}; 
            \end{tikz} & 2 & $\checkmark$\\
             \begin{tikz}
                \node [draw,fill=pink] {}; 
            \end{tikz} & 2 & $\text{\xmark}$\\
                 \begin{tikz}
                \node [draw,fill=white] {}; 
            \end{tikz} & \multicolumn{2}{ c}{causal mask}\\
                \bottomrule
            \end{tabular}
        };

    \end{tikzpicture}
\vspace{-2mm}
    \caption{
    Attention-style map to illustrate the chunkwise computations in GLA. The inter-chunk dependencies (in gray) 
    are not directly computed in the chunkwise form (only computed in the parallel form).
    The intra-chunk dependencies are modeled via secondary chunking/tiling where the inter-sub-chunk part (in orange) is computed by half-precision matmuls while the intra-sub-chunk part (in pink) is computed in full precision in log space. 
    }
    \label{fig:gla_tiling}

\end{figure}

\vspace{-2mm}
\subsection{Chunkwise Parallel Form of GLA}
\vspace{-2mm}
We derive a chunkwise form of GLA similar to the chunkwise form of basic linear attention (\S\ref{background:lin-chunkwise}). Here the intra-chunk operation implements the above parallel form at the chunk-level to obtain $\rmO^{\text{intra}}$. For inter-chunk, we have
\begin{align*}
\mathbf{\Lambda}_{iC+j} &= \frac{\vb_{iC+j}}{\vb_{iC}},  \mathbf{\Gamma}_{iC+j} = \frac{\vb_{(i+1)C}}{\vb_{iC+j}}, 
\bgamma_{i+1} = \frac{\vb_{(i+1)C}}{\vb_{iC}},  
 \\
\rmS_{[i+1]} &= \left(\bgamma_{i+1}^{\top}\mathbf{1}\right)  \odot \rmS_{[i]} + \left(\rmK_{[i+1]} \odot \mathbf \Gamma_{[i+1]} \right)^{\top}
\rmV_{[i+1]},  \\
\rmO^{\text{inter}}_{[i+1]} &= \left(\rmQ_{[i+1]} \odot \mathbf \Lambda_{[i+1]}\right) \rmS_{[i]}. 
\end{align*}
Intuitively, $\mathbf \Lambda_{[i+1]} $ encodes the cumulative decay from the start of a chunk which will be used to propagate the hidden states from the previous chunk $\rmS_{[i]}$, while   $\mathbf \Gamma_{[i+1]}$ encodes the decay to the end of a chunk which will be used to accumulate information to be added to the next hidden state $\rmS_{[i+1]}$.
\vspace{-2mm}
\subsection{Hardware-Efficient GLA}
\vspace{-2mm}
\label{subsec:flash-gla}
With the chunkwise form in hand, we can adapt the \textsc{FlashLinear Attention} algorithm presented in \S\ref{sec:algorithm} to the gated case. The adaptation additionally relies on two crucial techniques described below. 
We give high-level intuitions in this section
and defer the full algorithms to Alg. \ref{algo:gla-chunk-scan-1-fwd}-\ref{algo:gla-chunk-scan-2-bwd} of Appendix~\ref{sec:gla_algo}.

\vspace{-2mm}

\paragraph{Secondary-level chunking.}
Unlike in ordinary linear attention, the intra-chunk computations in GLA {cannot} leverage half-precision matmuls (and thus tensor cores) due to log space computations (Eq.~\ref{eq:log-semiring}). To make better use of tensor cores, we use secondary-level chunking scheme, where a chunk is further divided into sub-chunks (i.e., another level of tiling) in the spirit of classic tiling techniques~\cite{flashattention1}. The  attention-like matrix $\rmP \in \R^{L\times L}$ is then computed in a chunkwise manner, as illustrated in Figure~\ref{fig:gla_tiling}. 
 Concretely, the interactions  between sub-chunks are computed via half-precision matmuls,\footnote{To reduce notational clutter, here we use the notations from the first-level chunking to express the key idea. The actual implementation is done with secondary-level chunks.}
\begin{align*}
 \rmP_{[i][j]} &=  \Big(\rmQ_{[i]} \odot \bLambda_{[i]} \Big) \Big(\rmK_{[j]} \odot \bGamma_{[j]} \odot  
 \frac{\vb_{iC}}{\vb_{(j+1)C}} \Big)^{\intercal} \in \mathbb{R}^{C\times C}.
 \label{eq:gla-subchunk-compute-p}
\end{align*}
This corresponds to the orange tiles in Figure~\ref{fig:gla_tiling}. For the intra-sub-chunk part (pink tiles in Figure~\ref{fig:gla_tiling}) we have to resort to Eq.~\ref{eq:log-semiring} and perform the matmul in full precision for stability.
With this two-level tiling strategy, the total amount of non-half-precision matmul FLOPs are greatly reduced, thus leading to wallclock improvements. We provide the Pytorch-style pseudo-code in Listing~\ref{a} of Appendix~\ref{sec:gla_algo}.

\vspace{-2mm}
\paragraph{Memory-efficient  \textnormal{$\dbalpha_t$} computation.} 
Past work \citep[\S 3.1]{mao-2022-fine} has claimed that GLA-like models have to materialize the matrix-valued hidden states of size $L \times d \times d$ in HBM to compute all the gradients $\dbalpha_t$, since $\dbalpha_t =  (\rmS_{t-1} \odot \mathbf d\rmS_{t}) \mathbf{1}$. 
We instead give the following \emph{closed form} formula for $\dblogalpha_t$, 
\begin{align*}
    \vdlogb_t &= \vq_t \odot \vdq_t - \vk_t \odot \vdk_t,  \hspace{4mm}
   \dblogalpha_t  = \sum_{t \leq i \leq L} \vdlogb_i,
\end{align*}
which can be easily obtained by taking the derivative with respect to 
 Eq.~\ref{eq:log-semiring} (see Appendix~\ref{sec:gla_algo} for full derivation). $\vdq_t$ and $\vdk_t$ can be  computed  as in Alg.~\ref{algo:la-chunk-bwd}.
 
\vspace{-2mm}
\subsection{GLA Transformer}
\vspace{-2mm}
\label{sec:gla-full}
We generalize the GLA layer to the multi-head case. Given $H$ heads, we have the following for each head $h \in [1, H]$,
\begin{align*}
    &\rmS^h_{t} =  \left( \left(\balpha_{t}^h\right)^{\top} \mathbf{1} \right)
    \odot \rmS_{t-1}^h + \vk_t^{h \intercal} \, \vv^h_t   \in \mathbb{R}^{d'_k \times d'_v}, \\  %
& \vo^h_{t} = \vq_t^{h}\rmS_t^h    \in \mathbb{R}^{1 \times d'_v}, \\ &
\vo'_t = \operatorname{concat}(\lnorm(\vo^1_t), \dots, \lnorm(\vo^H_t)) \in \R^{1 \times d_v}, 
  \\
 &\vr_t = \operatorname{Swish}(\vx_t \mW_r + \vb_r)  \in \mathbb{R}^{1 \times d_v}, 
 \\ & \vy_t = (\vr_t \odot \vo'_t) \mW_{O}  \in \R^{1 \times d}. 
\end{align*}
Here we use separate key ($d_k$) and value ($d_v$) dimensions; $d'_k = d_k / H, d'_v = d_v / H$ are the per-head key/value dimensions. LayerNorm ($\lnorm$) is applied after the output of each head, while the output projection and output gating operate on the concatenation of head outputs \cite{sun2023retentive}. 

We then build up a Transformer-like model by interleaving multi-head GLA layers with feed-forward networks (FFN). Concretely, given layer $l$'s contextualized representation $\rmX^{(l)}$, we obtain  $\rmX^{(l+1)}$ via,
\begin{align*}
   &\rmY^{(l)} = \operatorname{GLA}(\lnorm(\rmX^{(l)})) + \rmX^{(l)} \\ 
   & \rmX^{(l+1)} = \operatorname{SwiGLU}(\lnorm(\rmY^{(l)})) + \rmX^{(l)},
\end{align*}
where the $\swiglu$ FFN layer~\citep{touvron2023llama} is,
\begin{align*}
    \operatorname{SwiGLU}(\rmZ) = (\operatorname{Swish}(\rmZ \mW_1) \odot \rmZ \mW_2) \mW_{3}.
\end{align*}

\begin{table*}[t!]
\centering
\small
\addtolength{\tabcolsep}{-2.5pt}    
\begin{tabular}{l l|cc|cccccc|c}
\toprule
&   & \textbf{Wiki.}  &  \textbf{LMB.} &  \textbf{LMB.} & \textbf{PIQA} &    \textbf{Hella.} & \textbf{Wino.} & \textbf{ARC-e} &  \textbf{ARC-c} &  \textbf{Avg.}  \\
\textbf{Scale} & \textbf{Model}  & ppl $\downarrow$  &  ppl $\downarrow$  &  acc $\uparrow$  & acc $\uparrow$ &   acc\_norm $\uparrow$  & acc $\uparrow$  & acc $\uparrow$ & acc\_norm $\uparrow$ &  $\uparrow$ \\
\midrule
\textit{340M Params} &  Transformer++ & \text{28.39} & 42.69 & \text{31.0} & 63.3 & 34.0 &	50.4 & 44.5	& \text{24.2} &  41.2  \\
\textit{15B Tokens} &   RetNet  & 32.33 &	49.19 &	28.6 &	63.5 & 33.5 &	\text{52.5} &	44.5 &	23.4 &	 41.0 \\
&   Mamba  & \text{28.39} & \text{39.66} & 30.6 & \text{65.0} & \text{35.4} & 50.1 & \text{46.3} & 23.6 & \text{41.8} \\ 
&   GLA   & 28.65 &	43.35 &	30.3 &	64.8  & 34.5 & 51.4  & 45.1	 & 22.7 &	 41.5\\
\midrule 
 \textit{1.3B Params} &  Transformer++ & \text{16.85} & \text{13.44} &  \text{48.9} & 70.8 & 49.6 & 53.6 & 56.0 & 26.5 & 50.9 \\
\textit{100B Tokens}&   RetNet  & 18.64 & 17.27 & 43.3 & 70.0 & 47.3 & 52.5 & 54.8 & 25.6 & 48.9 \\
&  Mamba  & 17.06 & 13.89 & 46.2 & \text{72.2} &  40.1 & \text{54.1} &  \text{59.0} &	\text{28.2} & 50.0 \\
&  GLA  & 17.22 & 14.47 & 46.9 & 71.8 & \text{49.8} & 53.9 & 57.2 & 26.6 & \text{51.0}  \\
\bottomrule
\end{tabular}
\addtolength{\tabcolsep}{2.5pt}    
\centering
\vspace{-2mm}
\caption{GLA Transformer results against Transformer++ \citep{touvron2023llama}, RetNet \citep{sun2023retentive}, and Mamba \citep{Gu2023MambaLS}. All models are trained on the same subset of the SlimPajama dataset with the Mistral tokenizer. The 340M/1.3B models are trained for 15B/100B tokens respectively. The individual task performance is via zero-shot. We report the main results on the same set of tasks reported by \citet{Gu2023MambaLS}. See Appendix~\ref{app:d} for results on other benchmarks, including 5-shot results. The last column shows the average over all benchmarks that use (normalized) accuracy as the metric.}
\vspace{-3mm}
\label{tab:main_results}
\end{table*}

\vspace{-2mm}
\paragraph{Parameter allocation.}  As presented, our GLA layer employs two additional matrices  for predicting $\balpha_t, \vr_t$ (i.e., $\mW_\alpha, \mW_r$) compared to a regular softmax attention layer. %
For parameter-efficiency, we use a low-rank parameterization 
\begin{align*}
   \balpha_t = \sigma((\vx_t \mW^1_\alpha \mW^2_\alpha + \vb_\alpha)))^{\frac{1}{\tau}}   \in \R^{1 \times d_k},
\end{align*}
where $\mW^1_\alpha \in \R^{d \times 16}$, $\mW^2_\alpha \in \R^{16 \times d_k}$, and $\tau = 16$ is a temperature term to encourage model to have a slower forgetting rate. 
We further set $d_k = \frac{d}{2}$ and $d_v = d$ and use  full-rank parameterizations for ($\mW_Q, \mW_K, \mW_V, \mW_O, \mW_r$). 
Ultimately, one GLA layer collectively needs (roughly) $4d^2$ parameters, as in regular softmax attention.

\vspace{-2mm}
\section{Empirical Study}
\vspace{-2mm}
\label{sec:experiments}

\subsection{Experimental Setup}
\vspace{-2mm}
Our main experiments are on language modeling, where we study whether  GLA  can perform competitively against a (i) strong Transformer baseline with modern architectural recipes and (ii) recent linear-time models. We use the SlimPajama dataset~\citep{cerebras2023slimpajama} and tokenize it using the Mistral tokenizer~\citep{jiang2023mistral}. The original dataset contains 627B tokens; we use a 100B subset.

\vspace{-2mm}
\paragraph{Baselines.} We evaluate  GLA  against three baselines: Transformer++ \citep{touvron2023llama}, RetNet \citep{sun2023retentive}, and Mamba \citep{Gu2023MambaLS}. 
Transformer++ is the LLaMA  architecture with Rotary Positional Embeddings~\citep{rope}, SWiGLU \citep{shazeer2020glu}, and RMSNorm~\citep{rmsnorm}; we also use SwiGLU in the RetNet to replace its original FFN for fair comparison. For Mamba, we use the open source code. All our baselines are trained for the exact same number of tokens on the same dataset for fair comparison.

\vspace{-2mm}
\paragraph{Training details.} We train all models from scratch at two scales: 340M and 1.3B.  All models are trained with AdamW~\citep{loshchilov2018fixing} using a maximum learning rate of 3e-4. The 340M models are trained on 15B tokens with a batch size of 0.5M tokens, while the 1.3B models are trained on 100B tokens with a batch size of 2M tokens. We use a cosine learning rate schedule with a warmup of 0.5B/1B tokens for the 340M/1.3B settings, respectively. The initial and final learning rates are 3e-5. We use a weight decay of 0.01, and gradient clipping of 1.0.

\vspace{-2mm}
\subsection{Main Results}
\vspace{-2mm}

In addition to perplexity (ppl) on Wikitext (Wiki.), we consider a wide range of downstream tasks covering common-sense reasoning and question-answering as was used in \citet{Gu2023MambaLS}: LAMBADA~\citep[LMB.; ][]{paperno2016lambada}, PiQA~\citep{bisk2020piqa}, HellaSwag~\citep[Hella.; ][]{zellers2019hellaswag}, WinoGrande~\citep[Wino.;][]{sakaguchi2021winogrande}, ARC-easy (ARC-e) and ARC-challenge (Arc-c) \citep{arc-ce}.
In Appendix~\ref{app:d}, we also include results on additional tasks: Copa~\citep{copa}, SciQA~\citep{SciQA2023}, OpenbookQA~\citep{openbookqa}, BoolQA~\citep{clark2019boolq}. We report perplexity (ppl) on WikiText and LAMBADA, accuracy normalized by length on
HellaSwag, ARC-challenge and OpenbookQA, and accuracy on the other tasks. All evaluations are performed using the LM evaluation harness~\citep{eval-harness}.

\definecolor{color1}{RGB}{116, 184, 22}
\definecolor{color2}{RGB}{77, 171, 247}
\definecolor{color3}{RGB}{99, 230, 190}
\definecolor{color4}{RGB}{132, 94, 247}
\definecolor{color5}{RGB}{250, 176, 5}
	
	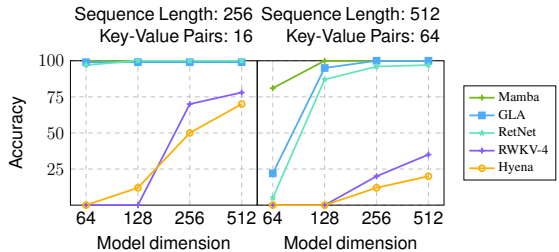
\begin{figure}[t] %
     \vspace{-4mm}
		\centering %
		\resizebox{0.9 \columnwidth}{!}{  %
			\begin{tikzpicture} %
			\scalefont{1.2} %
			\begin{axis}[
                name=plot1,
			sharp plot, %
        title style={align=right}, title={Sequence Length: 256\\Key-Value Pairs: 16},
			xmode=normal,%
			xlabel=Model dimension, %
			ylabel=Accuracy, %
			width=6cm, height=5cm,  %
			ymin=0, ymax=100,  %
                symbolic x coords={64,128,256,512},
			ytick={25,50,75,100}, %
			xlabel near ticks, %
			ylabel near ticks, %
			ymajorgrids=true, %
                xmajorgrids=true,
			grid style=dashed, %
			]
			\addplot+[very thick,mark=x,mark options={scale=0.5}, color=color1] plot coordinates { 
				(64,100)
				(128,100)
				(256,100)
				(512,100)
			};

			\addplot+[very thick, mark=square*, mark options={scale=1}, color=color2] plot coordinates {
				(64,99)
				(128,99)
				(256,99)
				(512,99)
			};

			\addplot+[very thick,mark=star,mark options={scale=1}, color=color3] plot coordinates {
				(64,97)
				(128,100)
				(256,100)
				(512,100)
			};
   			\addplot+[very thick,mark=+,mark options={scale=1}, color=color4] plot coordinates {
				(64,0)
				(128,0)
				(256,70)
				(512,78)
			};

			\addplot+[very thick,mark=o,mark options={scale=1}, color=color5] plot coordinates {
				(64,0)
				(128,12)
				(256,50)
				(512,70)
			};

			\end{axis}
			\begin{axis}[
                name=plot2,
                at={(plot1.south east) },
			sharp plot, %
        title style={align=right}, title={Sequence Length: 512\\Key-Value Pairs: 64},
			xmode=normal,%
			xlabel=Model dimension, %
			width=6cm, height=5cm,  %
			ymin=0, ymax=100,  %
                symbolic x coords={64,128,256,512},
   			ytick={0,25,50,75,100}, %
         yticklabels={,,},
			xlabel near ticks, %
			ymajorgrids=true, %
			xmajorgrids=true, %
			grid style=dashed, %
			legend style={at={(1.6,0.5)},anchor=east, legend cell align=left, font=\small}, 
			]
			
			\addplot+[very thick,mark=+,mark options={scale=1}, color=color1] plot coordinates { 
				(64,81)
				(128,100)
				(256,100)
				(512,100)
			};
			\addlegendentry{Mamba} 
			
			\addplot+[very thick, mark=square*, mark options={scale=1}, color=color2] plot coordinates {
				(64,22)
				(128,95)
				(256,100)
				(512,100)
			};
			\addlegendentry{GLA}

			\addplot+[very thick, mark=star, mark options={scale=1}, color=color3] plot coordinates {
				(64,5)
				(128,87)
				(256,96)
				(512,97)
			};
			\addlegendentry{RetNet}  

			\addplot+[very thick, mark=+, mark options={scale=1}, color=color4] plot coordinates {
				(64,0)
				(128,0)
				(256,20)
				(512,35)
			};
			\addlegendentry{RWKV-4}  

		\addplot+[very thick, mark=o, mark options={scale=1}, color=color5] plot coordinates {
				(64,0)
				(128,0)
				(256,12)
				(512,20)
			};
			\addlegendentry{Hyena}

			\end{axis}

			\end{tikzpicture}
		}
     \vspace{-3mm}
		\caption{Accuracy (\%) on the synthetic MQAR task.} 
		\label{fig:mqar} 

	\end{figure}
\definecolor{color1}{RGB}{77, 171, 247}
\definecolor{color2}{RGB}{132, 94, 247}
\definecolor{color3}{RGB}{255, 135, 135}
\definecolor{color4}{RGB}{11, 114, 133}
\definecolor{color5}{RGB}{59, 201, 219}
\definecolor{color6}{RGB}{132, 94, 247}

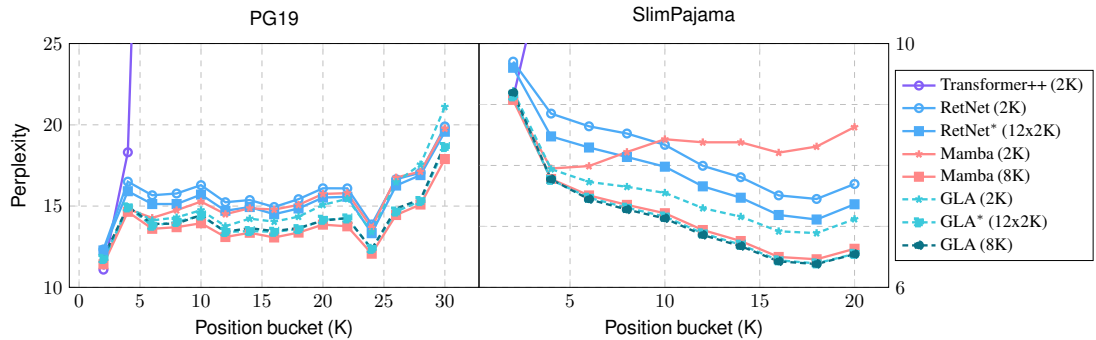
\begin{figure*}[t] %
     \vspace{-3mm}
    \centering %
    \resizebox{0.85\linewidth}{!}{  %
\begin{tikzpicture} %
        \scalefont{1.0} %
        \begin{axis}[
            name=plot1,
        sharp plot, %
        title style={align=right}, title={
        PG19}
        ,
			xmode=normal,%
			xlabel=Position bucket (K), %
			ylabel=Perplexity, %
			width=9cm, height=6cm,  %
			ymin=10,
               ymax=25,  %
			xlabel near ticks, %
			ylabel near ticks, %
			ymajorgrids=true, %
                xmajorgrids=true,
			grid style=dashed, %
			]

\addplot+[very thick,mark=o,mark options={scale=1}, color=color6] plot coordinates { 
(2, 11.09)
(4, 18.31)
(6, 64.2)
};

\addplot+[very thick,mark=o,mark options={scale=1}, color=color1] plot coordinates { 
(2, 12.334416871694545)
(4, 16.50338878264782)
(6, 15.665969802718664)
(8, 15.775098093939631)
(10, 16.28891258448325)
(12, 15.237502382790032)
(14, 15.371704482324168)
(16, 14.94472773435605)
(18, 15.426467447568902)
(20, 16.095521862334422)
(22, 16.089831898143757)
(24, 13.870173108390853)
(26, 16.652435886300694)
(28, 17.05460165670472)
(30, 19.899968051899997)
};

\addplot+[very thick, mark=square*, mark options={scale=1}, color=color1] plot coordinates {
(2, 12.272513995002527)
(4, 15.915017718544624)
(6, 15.12648400208903)
(8, 15.126682357304789)
(10, 15.707573736739828)
(12, 14.711491336539076)
(14, 14.941795590117444)
(16, 14.499628795410947)
(18, 14.852281588773375)
(20, 15.52356082374071)
(22, 15.57131917875397)
(24, 13.332259116220783)
(26, 16.26334764613957)
(28, 16.90620689149074)
(30, 19.575600800473826)	
};

\addplot+[very thick,mark=star,mark options={scale=1}, color=color3] plot coordinates {
(2, 11.513764074464719)
(4, 14.821545167608454)
(6, 14.27181440333812)
(8, 14.745159645587211)
(10, 15.277090611163125)
(12, 14.503881505445126)
(14, 14.867072874125807)
(16, 14.785802233682734)
(18, 15.075549715264373)
(20, 15.74728691109622)
(22, 15.796882231799712)
(24, 13.665838224478483)
(26, 16.75761700477185)
(28, 17.122188320649027)
(30, 19.772022829026763)			
};

\addplot+[very thick,mark=square*,mark options={scale=1}, color=color3] plot coordinates {
(2, 11.420266254807174)
(4, 14.647489438851771)
(6, 13.602399098526789)
(8, 13.702136965988363)
(10, 13.94407030138506)
(12, 13.096784138731092)
(14, 13.358225399949397)
(16, 13.069880104168242)
(18, 13.369938037195737)
(20, 13.859198521307533)
(22, 13.75584104547106)
(24, 12.071451636850792)
(26, 14.451484019606044)
(28, 15.095705584933034)
(30, 17.893014429053675)	
};

\addplot+[very thick,mark=star,mark options={scale=1}, color=color5] plot coordinates {
(2, 11.688854804956376)
(4, 14.943035357330759)
(6, 14.119485985291082)
(8, 14.300316332465233)
(10, 14.748137591710703)
(12, 13.762720187004728)
(14, 14.227475231817666)
(16, 14.04244932822711)
(18, 14.34961326894769)
(20, 15.074579287890959)
(22, 15.446429587912565)
(24, 13.412177117829918)
(26, 16.474816040581416)
(28, 17.556498322948684)
(30, 21.106791888790198)	
};

\addplot+[very thick,mark=square*,mark options={scale=1}, color=color5] plot coordinates {
(2, 11.7457580111041)
(4, 14.920393773306506)
(6, 13.791731057875367)
(8, 14.025104116895617)
(10, 14.419469470911702)
(12, 13.408896666345553)
(14, 13.464048125083734)
(16, 13.417051316705868)
(18, 13.58252067910424)
(20, 14.157938178864061)
(22, 14.239108084631486)
(24, 12.368099123010358)
(26, 14.665233740423941)
(28, 15.31096912196873)
(30, 18.632639168708696)
};

\addplot+[very thick, mark= mark options={scale=1},  color=color4] plot coordinates {
(2, 11.872848224059455)
(4, 14.972456535259505)
(6, 13.906079604037489)
(8, 13.917668674699033)
(10, 14.450326377207194)
(12, 13.422948137513114)
(14, 13.644676392922488)
(16, 13.46075819589147)
(18, 13.653410626666675)
(20, 14.103521696916971)
(22, 14.265809961807271)
(24, 12.348204946403571)
(26, 14.846425830412182)
(28, 15.410869769068253)
(30, 18.799468389849164)
};

\end{axis}

\begin{axis}[
            name=plot2,
            at={(plot1.south east) },
        sharp plot, %
        title style={align=right}, title={
        SlimPajama}
        ,
			xmode=normal,%
			xlabel=Position bucket (K), %
      ylabel near ticks, yticklabel pos=right,
			width=9cm, height=6cm,  %
   ymax=10,  %
   ymin=6,
			xlabel near ticks, %
			ylabel near ticks, %
			ymajorgrids=true, %
                xmajorgrids=true,
			grid style=dashed, %
   			legend style={at={(1.5,0.5)},anchor=east,font=\small}, 
                legend cell align=left]

			\addplot+[very thick,mark=o,mark options={scale=1}, color=color6] plot coordinates { 
(2, 9.11)
(4, 11.77)
};			
\addlegendentry{Transformer++ (2K)} ;

			\addplot+[very thick,mark=o,mark options={scale=1}, color=color1] plot coordinates { 
(2, 9.703324579039524)
(4, 8.851140172053045)
(6, 8.64224978739121)
(8, 8.523894072608067)
(10, 8.337787326052817)
(12, 7.993996058951587)
(14, 7.807283256053823)
(16, 7.508849714252663)
(18, 7.450385184846358)
(20, 7.696067587961978)			};
	\addlegendentry{RetNet (2K)};  

			\addplot+[very thick, mark=square*, mark options={scale=1}, color=color1] plot coordinates {
(2, 9.60288660871554)
(4, 8.47661116643274)
(6, 8.294459081541115)
(8, 8.138966746337)
(10, 7.979737282823978)
(12, 7.65726433276238)
(14, 7.470669790601427)
(16, 7.187401744904839)
(18, 7.1132779884136905)
(20, 7.3631378999335295)		};
	\addlegendentry{RetNet$^*$ (12x2K) };  

			\addplot+[very thick,mark=star,mark options={scale=1}, color=color3] plot coordinates {
(2, 9.07)
(4, 7.95)
(6, 7.99)
(8, 8.22)
(10, 8.43)
(12, 8.38)
(14, 8.38)
(16, 8.21)
(18, 8.31)
(20, 8.63)		};
	\addlegendentry{Mamba (2K) };  

   			\addplot+[very thick,mark=square*,mark options={scale=1}, color=color3] plot coordinates {
(2, 9.074799787988505)
(4, 7.7780675203930025)
(6, 7.508237473960171)
(8, 7.3529524666619945)
(10, 7.216372863755205)
(12, 6.940692359274163)
(14, 6.760323233107172)
(16, 6.49872027703192)
(18, 6.460961900906283)
(20, 6.6330331428715725)	};
	\addlegendentry{Mamba (8K) };  

\addplot+[very thick,mark=star,mark options={scale=1}, color=color5] plot coordinates {
(2, 9.22)
(4, 7.94)
(6, 7.73)
(8, 7.65)
(10, 7.55)
(12, 7.3)
(14, 7.16)
(16, 6.92)
(18, 6.89)
(20, 7.12)		};
\addlegendentry{GLA (2K) };  

\addplot+[very thick,mark=square*,mark options={scale=1}, color=color5] plot coordinates {
  (2, 9.134076157537933)
(4, 7.763725571624473)
(6, 7.470406185573722)
(8, 7.307020316318466)
(10, 7.162166118009677)
(12, 6.8859825063022155)
(14, 6.69981277659273)
(16, 6.443211499058021)
(18, 6.393198065692631)
(20, 6.54966772918719)
};
\addlegendentry{GLA$^{*}$ (12x2K) };  
\addplot+[very thick, mark options={scale=1},  color=color4] plot coordinates {
(2, 9.195530723286662)
(4, 7.773488412387971)
(6, 7.445253416398666)
(8, 7.276337205511129)
(10, 7.131082988866674)
(12, 6.858498031943685)
(14, 6.67630503682912)
(16, 6.426108888086446)
(18, 6.382840753376257)
(20, 6.543979086587886)
};
\addlegendentry{GLA (8K) };  
\end{axis}
\end{tikzpicture}
}
 \vspace{-3mm}
\caption{Length extrapolation on the test set of SlimPajama and  PG19. We pretrain 1.3B models from scratch on SlimPajama for 100B tokens with different training length. $^{\ast}$ indicates models using truncated BPTT with over 12 segments that are each of 2K-length.}
		\label{fig:length_extrapolate} 
     \vspace{-4.5mm}
	\end{figure*}

Our main results are shown in Table~\ref{tab:main_results}. Compared to RetNet which uses a data-independent decay rate, the GLA Transformer with data-dependent gates shows improved results on all tasks. Both GLA Transformer and Mamba show comparable performance to Transformer++. 
\vspace{-2mm}

\paragraph{Recall-intensive tasks.} While subquadratic models can achieve competitive language modeling performance to Transformers, \citet{Arora2024SimpleLA} show that they  lag behind softmax attention in recall-intensive tasks. We next  evaluate GLA on real and synthetic tasks that focus on recall. 

The synthetic MQAR task \cite{zoology} is a more challenging multi-query version of the induction head task \cite{h3} in which a model has to recall the token following a query token multiple times.   We follow 
\citet{zoology}'s experimental setting and compare GLA against recent subquadractic models, including RetNet \cite{sun2023retentive}, Mamba \cite{Gu2023MambaLS}, Hyena \cite{hyena} and  RWKV-4 \cite{rwkv}. For RetNet and GLA the number of heads is set to 2; for other models we follow the default settings in \citet{zoology}. The results are shown in Figure~\ref{fig:mqar}. Standard quadratic attention achieves perfect scores in all settings and is thus omitted. We find that models with matrix-valued hidden states (i.e., Mamba/RetNet/GLA) outperform Hyena/RWKV, and our GLA outperforms RetNet, confirming the benefits of using data-dependent gates.

Following \citet{Arora2024SimpleLA}, we also test our models on three real recall-intensive tasks: FDA \cite{arora_language_2023}, SWDE \cite{lockard_openceres_2019}, and SQUAD \cite{rajpurkar_know_2018}. These tasks focus on information extraction or reading comprehension. As illustrated in Table~\ref{tab:recall-intensive}, subquadratic models significantly underperform Transformers on the FDA and SWDE, both of which are information extraction tasks. However, GLA outperforms other subquadractic models, likely due to its larger recurrent state (compared to Mamba) and selection mechanism (compared to RetNet).

\begin{table}

\centering
\small 
        \begin{tabular}{l lccc}
        \toprule[0.5mm]
  \textbf{Scale} &   \textbf{Model}  &  \textbf{FDA} & \textbf{SWDE} & \textbf{SQUAD} \\
      \bottomrule
 \textit{340M Params}  &   Transformer++ &  21.4 & 42.2 & 22.1 \\
 \textit{15B Tokens}  &   RetNet & 2.9 & 13.3 & 27.6 \\
   &   Mamba & 2.1 & 12.4 & 23.0 \\
   &   GLA  & 8.1 & 18.6 & 27.2 \\      
      \midrule
 \textit{1.3B Params}   &   Transformer++ & 27.4 & 66.6 & 31.5 \\
 \textit{100B Tokens}   &   RetNet & 14.3 & 42.8 & 34.7 \\
   &   Mamba &  6.2 & 41.4 & 35.2 \\
   &   GLA  & 19.9 &  50.6 & 42.6 \\      
    \bottomrule
    \end{tabular}
    \vspace{-2mm}
    \caption{Comparison of different models in three recall-intensive tasks tested in \citet{Arora2024SimpleLA}. Higher is better for all tasks. 
    }
    \label{tab:recall-intensive}
\end{table}

\vspace{-2mm}
\paragraph{Long sequence training and length extrapolation.} 
One advantage of linear attention models is that they allow for efficient long sequence training in linear time.
To showcase this feature, we consider two training settings: (i) direct training on 8K-length contexts, (ii) training on 24K-length contexts through {truncated backpropagation through time} (TBPP) over 2K-length segments.\footnote{We split a 24K input sequence into 12 segments. The final state of the previous segment is used as the initial state for the current segment.} In the latter case the gradients are not back-propagated across segments, and hence this approach has minimal overhead comparable to the standard 2K-length training strategy (where the initial hidden state is always set to zero). We pretrain 1.3B Mamba, RetNet, and GLA models on SlimPajama for 100B tokens on these settings and test them on both SlimPajama test set and PG19 \citep{pg19} test set. 

Figure \ref{fig:length_extrapolate} shows the perplexities of the tokens calculated in different position groups.  For models trained on 2K-length contexts, GLA extrapolates better than Mamba/RetNet in most position buckets on the PG19 test set; Mamba struggles to extrapolate beyond 4K, while GLA/RetNet can generalize to 18K on the Slimpajama test set.
Transformers cannot extrapolate beyond training length, which is a known failure mode.\footnote
{Although there are positional encoding schemes that enable better length extrapolation, these methods still have difficulty generalizing significantly beyond context lengths seen during training~\citep{alibi2021,xpos,fire2024}.} Pretraining in a long sequence consistently improves perplexities for all three models. We found marginal perplexity difference in the two settings for GLA, indicating that TBPTT might be a more economic approach to long-sequence training. Mamba benefits significantly from 8K-length training, and it performs similarly as GLA in the same training setting.
\vspace{-2mm}
\paragraph{Ablations.}
We conduct a small-scale ablation study by training the 340M GLA variants for 7B tokens. We investigate (i) the importance of having both \emph{fine-grained} and \emph{data-dependent} gating  and (ii) the influence of head dimension size. The results are shown in Table~\ref{tab:ablation}. For (i), we find that while data dependent scalar gates substantially improve upon RetNet, a finer-grained gating mechanism is still necessary. For (ii) we tune the number of heads to vary head dimensions, where by default GLA uses 4 heads. Increasing it to 8 (\text{i.e., smaller head dimension}) leads to relatively large perplexity degradation; reducing it to 1 (i.e., larger head dimension) actually performs best, but results in only marginal improvement while requiring much higher GPU memory. We thus choose 4 heads for our experiments.
\begin{table}

\centering
\small 
        \begin{tabular}{ll}
        \toprule[0.5mm]
     Model variants  &   Training ppl. \\
      \bottomrule
        {GLA Transformer} (4 heads)  & 14.77 \\
\hspace{2mm} {No gate} (i.e., Linear Attention) & 23.21\\
\hspace{2mm} {\emph{Data independent} scalar decay} (i.e., RetNet) & 16.55 \\
\hspace{2mm} {\emph{Data dependent} scalar gate} & 15.56 \\
\hspace{2mm} {Small head dimension (8 heads)} & 15.29 \\
\hspace{2mm} {Large head dimension (1 head)} &  14.61 \\
        \bottomrule
        \end{tabular}
    \vspace{-2mm}
    \caption{Ablation study results on the 340M model trained for 7B tokens. We evaluate the model variants via the average perplexity of the last 200 training steps. 
    }
        \label{tab:ablation}

\end{table}

\vspace{-2mm}
\subsection{Training Efficiency}
\vspace{-2mm}

\begin{figure}[t] 
\vspace{-3mm}
		\centering 
		\resizebox{1\columnwidth}{!}{  
\begin{tikzpicture}
\scalefont{1.5} %
\begin{axis}[
    name=plot1,
    title={Training throughput},
        xlabel={Training length/Batch size},
    ylabel={Tokens per second (Kt/s)},
    major x tick style=transparent,
    ybar=2*\pgflinewidth,
    bar width=8pt,
    x tick label style={rotate=0, anchor=center},
    symbolic x coords={2048/8,4096/4,8192/2,16284/1},
    xticklabel style={yshift=-2mm}, 
    ymin=0,
    xtick=data,
    tick label style={font=\large},
    label style={font=\large},
    legend style={at={(7,-1.6)},   
                anchor=north,legend columns=3,
                column sep=0.3cm,
                font=\normalsize,
                row sep=-0.15cm
                },     
    legend cell align=left,
    ymajorgrids=true,
    grid style=dashed,
    enlarge x limits=0.25,
    point meta=explicit symbolic,
    legend to name={mylegend},
]
\addplot[draw=bred,thick,fill=bred,fill opacity=0.6] coordinates { 
  (2048/8, 51.3)[\textcolor{bred!60}]
  (4096/4, 46.7)[\textcolor{bred!60}]
  (8192/2, 38.7)[\textcolor{bred!60}]
   (16284/1, 29.1)[\textcolor{bred!60}]
};

\addplot[draw=bblue,thick,fill=bblue,fill opacity=0.6] coordinates {
  (2048/8, 22.8)[\textcolor{bblue!60}{1408}]
  (4096/4, 22.8)[\textcolor{bblue!60}{352}]
  (8192/2, 22.8)[\textcolor{bblue!60}{1056}]
  (16284/1, 26.0)[\textcolor{bblue!60}{1056}]
};

\addplot[draw=bcyan,thick,fill=bcyan,fill opacity=0.6] coordinates {
  (2048/8, 43.8)[\textcolor{bcyan!60}{1056}]
  (4096/4, 43.5)[\textcolor{bcyan!60}{352}]
  (8192/2, 43.2)[\textcolor{bcyan!60}{1056}]
  (16284/1, 41.1)[\textcolor{bcyan!60}{1056}]
  };

\legend{Transformer++, Mamba, GLA}

\end{axis}
\begin{axis}[
 at={(plot1.south east) },
    title={GPU memory usage},
    xlabel={Training length/Batch size},
    ylabel near ticks, yticklabel pos=right,
    ylabel={Gigabyte (GB)},
    major x tick style=transparent,
    ybar=2*\pgflinewidth,
    bar width=8pt,
    x tick label style={rotate=0, anchor=center},
    symbolic x coords={2048/8, 4096/4, 8192/2, 16284/1},
    xticklabel style={yshift=-2mm},
    xtick=data,
    ymin=0,
      tick label style={font=\large},
  label style={font=\large},
    ymajorgrids=true,
    grid style=dashed,
    enlarge x limits=0.25,
    point meta=explicit symbolic
]
\addplot[draw=bred,thick,fill=bred,fill opacity=0.6] coordinates { 
  (2048/8,33)[\textcolor{bred!60}]
  (4096/4, 33)[\textcolor{bred!60}]
  (8192/2, 33)[\textcolor{bred!60}]
   (16284/1, 33)[\textcolor{bred!60}]
};

\addplot[draw=bblue,thick,fill=bblue,fill opacity=0.6] coordinates {
  (2048/8, 36.0)[\textcolor{bblue!60}{1408}]
  (4096/4, 36.0)[\textcolor{bblue!60}{352}]
  (8192/2, 36.0)[\textcolor{bblue!60}{1056}]
  (16284/1, 36.0)[\textcolor{bblue!60}{1056}]
};

\addplot[draw=bcyan,thick,fill=bcyan,fill opacity=0.6] coordinates {
  (2048/8, 37.0)[\textcolor{bcyan!60}{1056}]
  (4096/4, 37.0)[\textcolor{bcyan!60}{352}]
  (8192/2, 37.0)[\textcolor{bcyan!60}{1056}]
  (16284/1, 37.0)[\textcolor{bcyan!60}{1056}]
  };
\end{axis}
\ref{mylegend}

\end{tikzpicture}
		}
     \vspace{-6mm}
		\caption{Training throughput and  memory footprint on an H100.} 
\label{fig:tps}
	\end{figure}
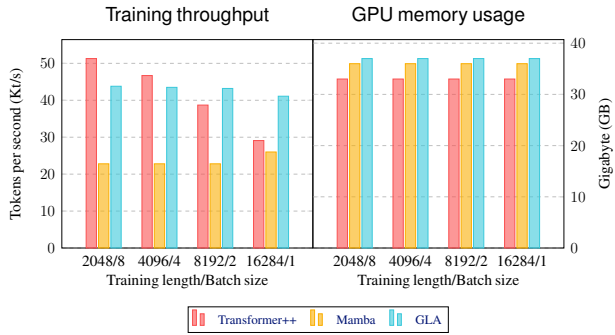

Fig.~\ref{fig:tps} shows the throughput and memory usage as a function of the sequence length and batch size for the different 1.3B models on a single H100 GPU.\footnote{We use the official implementation for Mamba, the fused version of SwiGLU  for Transformer++ and GLA, and FlashAttention-2 for Transformer++.} Here GLA adopts the materialization version of \textsc{FlashLinearAttention} with recomputation of hidden state (\S\ref{sec:fla}). All models have linear space complexity, and the total GPU footprint difference among them is minimal. In terms of training throughput, Mamba  lags behind Transformer++ and GLA, with GLA shows greater advantages in training lengths beyond 4096.

\vspace{-2mm}
\subsection{Limitations \& Future Work}
\vspace{-2mm}
While our experiments with the GLA Transformer were on a respectable scale, we were unable to perform larger-scale experiments due to limited compute resources.
Although it is unclear at this point how GLA would scale to even larger models/datasets, we anticipate that training efficiency of GLA become even more favorable compared to Mamba at larger scales. Specifically, when scaled to larger sizes (e.g., $>7$B), GLA can be more efficient than Mamba because of better use of tensor cores and GLA's compatibility with tensor parallelism.\footnote{In particular, since Mamba is not a  multi-head model it is not as amenable to tensor parallelism.} Insofar as we are interested in leveraging the efficiency of linear attention, it would be interesting to apply GLA to other modalities (especially modalities with long-range dependencies), in line with recent work on applying state-of-the-art state-space models to other types of data \citep[][\textit{inter alia}]{Yan2023DiffusionMW, zhu2024vision,ma2024u,liu2024vmamba,xing2024segmamba,wang2024graph,wang2024mambabyte,yang2024vivim}.

\vspace{-2mm}
\section{Related Work}
\label{sec:related}
\vspace{-2mm}
We briefly discuss related work here and give an extended discussion of the related work in Appendix~\ref{appdx:extended_rw}.

Traditional RNNs are difficult to scale  due to the nonlinear dependencies between the  hidden states and expensive matmul-based sequential hidden state updates. Linear RNNs/State-Space Models (SSMs)/Transformers eliminate nonlinear dependencies, making training parallelizable along the temporal dimension \citep{parallel-martin, s4, s5}. Such models have been the focus of much recent work as a competitive sub-quadratic alternative to the Transformer architecture \citep{rwkv, Gu2023MambaLS,HGRN,qin2023scaling,sun2023retentive,pretrain_wo_attn}.

Data-dependent decay rates have always been regarded important for  RNNs  \citep{DBLP:journals/neco/GersSC00,unreasonable-forget-gate}. Typical forget gate values depend on both the previous hidden state and the current input. However \citet{parallel-martin} suggest that forget gate values should depend solely on the current inputs to enable parallel training. This simple strategy has been shown to be effective in moderate-scale experiments conducted by HGRN \citep{qin2023scaling}.  RWKV-v6 \citep{peng2024eagle} and Mamba \citep{Gu2023MambaLS} also use data-dependent decay rates that are reminiscent of forget gates. In the context of linear Transformers, \citet{peng2021random} employ a coarse-grained position-wise forget gate, while \citet{mao-2022-fine} and \citet{gatedloop} use a more fine-grained forget gate. 

RNNs rely on fixed-dimensional hidden states to encode their entire history. The hidden state dimension  serves as a proxy for memory capacity and thus significantly influences their expressive power. 
Linear Transformers expand the hidden dimension of RNNs via the outer-product parameterization, as discussed \S\ref{subsec:background-lin}. Linear SSMs on the other hand expand their hidden dimension via a single-input-single-output (SISO) strategy. 
Without data-dependent SSM parameters, this can be done efficiently during training via the Fast Fourier Transform (FFT). However, with data-dependent SSM parameters, FFT-based training is not possible, and thus \citet{Gu2023MambaLS} implements a custom CUDA kernel to train a selective state-space model using the parallel scan algorithm \citep{s5}. To fit all the hidden states into SRAM, they can only afford an expansion rate up to 16. In contrast our hardware-aware training algorithm provides an alternative, efficient approach for expanding the hidden dimension to a wider range, which we have shown useful in recall-intensive tasks.

\vspace{-2mm}

\vspace{-2mm}
\section{Conclusion}
\vspace{-2mm}
We propose an efficient algorithm for training linear attention Transformers with data-dependent gating mechanisms. Our algorithm makes it possible to balance FLOPs against parallellism, while still allowing for the use of half-precision matmuls which can take advantage of tensor core units on modern GPUs. Experiments on language modeling demonstrate that gated linear attention Transformers can perform respectably compared to strong baselines.
\label{sec:conc}

\section*{Impact Statement}

This paper aims to improve the training efficiency of a new model family of (gated) linear attention models. The efficiency advantage of such models might help democratize access of language models. On the other hand, whether such new architectures would affect known issues such as biased and harmful outputs of language models remains an unexplored research question.

\section*{Acknowledgments}

This work was supported by MIT-IBM Watson AI Lab. We thank Yutao Sun, Zhen Qin, Li Dong, Xinyu Yang, Jiacheng You, Huanqi Cao, Yu Zhang, and Shida Wang for their insightful discussions. We also thank Yu Zhang, Fares Obeid, Daniel Goldstein, and Liliang Ren for their proofreading. Special thanks to Yu Zhang for contributing to the \textsc{FlashLinearAttention} library.

\bibliography{main}
\bibliographystyle{icml2024}
\newpage
\appendix

\onecolumn

\section{Extended Related Work }
\label{appdx:extended_rw}
\vspace{2mm}

\subsection{Linear Attention}

\paragraph{Feature map $\phi$.} Linear attention mechanisms~\citep{katharopoulos2020transformers} replace $\exp(\vq_t \vk_i^\intercal)$ with a kernel $k(\vx, \vy)$ having an associated feature map $\phi$ (i.e., $k(\vx, \vy) = \langle\phi(\vx), \phi(\vy)\rangle$) where $\phi \in \R^{d_{\text{key}}}\rightarrow \R^{d_{\text{dot}}}$.
$\phi$ often consists of two parts: $\phi = \phi_0 \circ \phi_1$. $\phi_1$ could be a linear map made up by random samples \cite{peng2021random, performer}, learnable MLPs \cite{kasai-etal-2021-finetuning, Hedgehog, polysketch} or simply an identity map \cite{mao-2022-fine}. $\phi_2$ is often an element-wise (activation) function that makes the resulting $\phi$ a positive feature map, such as $1+\operatorname{elu}$ \cite{katharopoulos2020transformers}, $\operatorname{ReLU}$ \cite{kasai-etal-2021-finetuning}, $\exp(\cdot)$ \cite{Hedgehog, performer}. Some work \cite{qin2023scaling, sun2023retentive, mao-2022-fine} suggests that a positive feature map might not be necessary.

Our work follows \citet{sun2023retentive} and \citet{mao-2022-fine} by using an identity map $\phi=\mathbf{I}$.  Recent work suggests that non-identity feature maps such as scaled element-wise exponential map  \cite{log-normal, Hedgehog}
and higher-order polynomial map \cite{Arora2024SimpleLA,polysketch} work well empirically. We leave the exploration of integrating other types of feature map into GLA to future work.

\paragraph{Attention spikiness.} Linear attention suffers from the ``attention dilution'' issue \citep{qin2022devil}, where the attention distribution is too uniform (i.e., high entropy) to concentrate on relevant tokens. \citet{qin2022devil} propose adding local attention layers to focus more on adjacent tokens, a method adopted in \citep{VQ-Transformer, log-normal, orthogonal_memory} and proven crucial for performance. 
Recent work finds that a scaled element-wise exponential map---i.e., $\phi(\mathbf{x})= \mathbf{\exp}(t\cdot \mathbf{x})$ with $t \ge 2$---helps to concentrate attention \cite{log-normal, Hedgehog}. \citet{Hedgehog} also find that higher-order polynomial kernels induce low-entropy and spiky attention distribution, partially explaining the empirical success of Based Linear Attention \cite{Arora2024SimpleLA} and PolySketchFormer \cite{polysketch}.

\paragraph{Memory capacity.} 
 Linear attention has bounded memory size \cite{peng-etal-2022-abc} while softmax attention enjoys unbounded memory\cite{Oren2024TransformersAM}. We believe that increasing the memory size efficiently and utilizing memory effectively are the keys to bridging the performance gap between linear attention and softmax attention. To increase memory size, it is shown that  directly increasing $d_{\operatorname{key}}$ is effective \citep{sun2023retentive, mao-2022-fine, zhang-cai-2022-linearizing}; however, the total parameters are hard to control with the increase of $d_{\operatorname{key}}$.
 Parameter-efficient methods often keep $d_{\text{key}}$ intact and increase $d_{\text{dot}}$ instead. Higher order polynomial kernels with order $p\ge2$ map $d_{\text{key}}$ to a much higher  $d_{\text{dot}} = O(d_\text{key}^p)$ \citep{zoology, polysketch}.  \citet{linear-xmr-fastweight} propose the Deterministic Parameter-Free Projection (DPFP), while \citet{recurrent_linear_xfmr} use parameterized outer product to expand  $d_{\text{dot}}$ in a parameter-efficient/free manner. 
 
 For better memory utilization, \citet{linear-xmr-fastweight} use the delta rule to edit the memory dynamically. However, this is shown to underperform the gating mechanism \cite{mao-2022-fine}, which is a classic method to erase irrelevant historical information in gated RNNs. Recently, \citet{orthogonal_memory} enforce orthogonality of memory vectors to potentially increase utiliziation.

 \paragraph{Linear attention with decay or gates.} \citet{peng2021random} use position-wise scalar gates for incorporating recency bias into linear attention, and has been revisited in recent work \cite{mamba2, beck2024xlstm, Sun2024YouOC},
 while \citet{mao-2022-fine, recurrent_linear_xfmr} use matrix-valued gates (obtained by the outer product) for more fine-grained memory control.

Scalar decays can be easily incorporated into chunkwise linear attention for training efficiency \citep{sun2023retentive, lightning2}. With matrix-valued gates, the training efficiency becomes much more challenging. 
 Both \citet{mao-2022-fine} and \citet{gatedloop}'s training algorithms involve materializing hidden states of all steps in HBM, which  suffers from high I/O costs. Moreover, both approaches cannot take advantage  of tensor cores. Our hardware-efficient training algorithm reduces or eliminates materialization and enables usage of tensor cores.

\paragraph{I/O-aware chunkwise linear attention.} 

The chunkwise form of  linear attention is well-known in the literature.  \citet{GAU} first propose the chunkwise linear attention form, arguing that the training algorithm of \citet{katharopoulos2020transformers} is slow in practice. 
\citet{sun2023retentive} and \citet{lightning2} generalize this form to linear attention with exponential decay (or ALiBi).  \citet{polysketch,VQ-Transformer} also derive similar chunkwise forms.

 However, most chunkwise linear attention is not I/O-aware. To the best of our knowledge, only \textsc{LightningAttention2} \cite{lightning2} (concurrent to our work) is I/O aware, and it is very similar to the non-materialization version of our \textsc{FlashLinearAttention}.  We additionally propose a materialization version, which leverages sequence-level parallelism and thus allows for higher training throughput at the cost of a slightly increasing memory footprint.

\paragraph{Other subquadratic models.} Besides the Linear attention Transformer \cite{katharopoulos2020transformers, linear-xmr-fastweight} discussed in this work, previous studies have explored sparsifying attention with either a predefined fixed pattern \cite{child2019generating, beltagy2020longformer, zaheer2020big} or a context-aware learnable pattern \cite{roy2020efficient, kitaev2020reformer, ren2023sparse} for sequence modeling with subquadratic complexity in the sequence length dimension. Leveraging convolutions for efficient sequence modeling has also been studied in works such as Dynamic Convolution \cite{wu2019pay}, Long Convolution \cite{fu2023simple, qin2023toeplitz, hyena, massaroli2023laughing, li2023what, romero2021ckconv}, and State Space Models \cite{gu2021efficiently, gupta2022diagonal, gu2021combining, hasani2022liquid, s5, ma2023mega}.

\subsection{Sequence parallelism}

The chunk-wise parallel form of linear Transformers resembles the two-stage parallel prefix sum (or parallel scan) algorithm \citep{Blelloch1990PrefixSA}, which also combine chunk-wise parallel computations with inter-chunk communication \citep{Chaurasia2015CompilingHP}.  It also resembles sequence parallelism used for accelerating attention-based Transformers  \citep{li-etal-2023-sequence}, which has recently received  much attention  for long-sequence modeling \citep{Liu2023RingAW, Li2023LightSeqSL, Brandon2023StripedAF}. Sequence-level parallelism also constitutes the main improvement of FlashAttention-2 \citep{flashattention2} over FlashAttention-1 \citep{flashattention1}. The main differences between these works are that (i) the chunk-level parallel form of linear Transformer needs only a single pass due to the linear complexity, while the sequence parallelism in Transformers needs $L/C$ passes (i.e., left-to-right scan of key/value blocks for each query block) due to the inherent quadratic complexity, and (ii) the order of matrix multiplications is different. We also note that chunkwise linear attention could greatly reduce the communication cost between devices in the distributed training setting compared to softmax attention, which could open the door for extremely long sequence training.

\begin{algorithm}[hb]
\scriptsize
\caption{\textsc{FlashLinearAttention}: Backward Pass}
\label{algo:la-chunk-bwd}
\begin{algorithmic}
    \Require $\rmQ, \rmK, \rmV, \rmO, \rmdO \in \R^{L \times d}$, chunk size $C \in [L]$,  \texttt{materialize} $\in$ \{\texttt{True,False}\}, $\rmS \in \R^{\frac{L}{C} \times d \times d}$ \Comment{$\rmS$ is available when \texttt{materialize} is True}
 \State{Initialize $\rmdS = \bm{0} \in \mathbb R^{d\times d} $ on SRAM}
         \State On chip, construct causal mask $\rmM\in \R^{C\times C}$ %
\If{\texttt{materialize}} \Comment{the materialization version}
\color{black}
    \For{$n \gets N, 1$}   \Comment{in reverse order}
        \State Store $\rmdS$ in HBM as $\rmdS_{[n]}$
    \color{black}
        \State{Load $\rmQ_{[n]}, \rmdO_{[n]}  \in \R^{C \times d}$ from HBM to SRAM.}
        \State On chip, compute $\rmdS = \rmdS + \rmQ_{[n]}^\intercal \rmdO_{[n]}$
        \color{orange}
            \color{black}
      \EndFor
      \ParFor{$n \gets 1, N$}  
        \State{Load $\rmQ_{[n]}, \rmK_{[n]},\rmV_{[n]}, \rmdO_{[n]} \in \R^{C \times d}$ from HBM to  SRAM.}
        \State{Load $ \rmS_{[n]}$, $\rmdS_{[n]} \in \R^{d \times d}$ from HBM to SRAM.}

   \State On chip:  $\mathbf{dQ} = \mathbf{dO}_{[n]} \rmS_{[n]}^{\top}  + (\mathbf{dO}_{[n]} \rmV_{[n]}^{\top} \odot \mathbf{M}) \mathbf{K}_{[n]}$.
   \State On chip:   $\mathbf{dK} =  \mathbf{V}_{[n]} \mathbf{dS}_{[n]}^{\top} +  (\mathbf{V}_{[n]}\mathbf{dO}_{[n]}^{\top} \odot \mathbf{M}^{\top}) \mathbf{Q}_{[n]}$
   \State On chip: $\mathbf{dV} =  \mathbf{K}_{[n]} \mathbf{dS}_{[n]} + (\rmQ_{[n]}\rmK_{[n]}^{\top} \odot \mathbf{M})^{\top} \mathbf{dO}_{[n]}$
    \State Write $ \mathbf {dQ}, \mathbf{dK}, \mathbf{dV}$ to HBM as $\mathbf{dQ}_{[n]}, \mathbf{dK}_{[n]}, \mathbf{dV}_{[n]}$
      \EndParFor
      \color{black}
\Else  \Comment{the non-materialization version}
\State Initial $\rmS = \bm{0} \in \mathbb R^{d\times d} $ on SRAM
\For{$n \gets 1, N$}    \Comment{hidden state recomputation}
\State Load $\rmK_{[n]}, \rmV_{[n]}, \rmdO_{[n]}   \in \R^{C \times d}$ from HBM to SRAM.
\State On chip: $\rmdQ = \mathbf{dO}_{[n]} \rmS^{\top}  + (\mathbf{dO}_{[n]} \rmV_{[n]}^{\top} \odot \mathbf{M}) \mathbf{K}_{[n]}$ 
\State On chip: $\rmS = \rmS + \rmK_{[n]}^{\top}\rmV_{[n]}$
\EndFor
\color{black}
    \For{$n \gets N, 1$}   \Comment{in reverse order}
    \color{black}
        \State{Load $\rmQ_{[n]},\rmK_{[n]}, \rmV_{[n]}, \rmdO_{[n]}  \in \R^{C \times d}$ from HBM to SRAM.}
        \State On chip, compute $\rmdS = \rmdS + \rmQ_{[n]}^\intercal \rmdO_{[n]}$
       \State On chip:  $\mathbf{dQ} = \mathbf{dO}_{[n]} \rmS_{[n]}^{\top}  + (\mathbf{dO}_{[n]} \rmV_{[n]}^{\top} \odot \mathbf{M}) \mathbf{K}_{[n]}$.
       \State On chip:   $\mathbf{dK} = \mathbf{V}_{[n]}\mathbf{dS}_{[n]}^{\top}  +  (\mathbf{V}_{[n]}\mathbf{dO}_{[n]}^{\top} \odot \mathbf{M}^{\top}) \mathbf{Q}_{[n]}$
       \State On chip: $\mathbf{dV} =  \mathbf{K}_{[n]} \mathbf{dS}_{[n]} + (\rmQ_{[n]}\rmK_{[n]}^{\top} \odot \mathbf{M})^{\top} \mathbf{dO}_{[n]}$
           \State Write $ \mathbf {dQ}, \mathbf{dK}, \mathbf{dV}$ to HBM as $\mathbf{dQ}_{[n]}, \mathbf{dK}_{[n]}, \mathbf{dV}_{[n]}$
            \color{black}
      \EndFor
\EndIf 
  \State \Return $\rmdQ = \{ \rmdQ_{[1]} \dots \rmdQ_{[N]} \}$, $\rmdK=\{ \rmdK_{[1]} \dots \rmdK_{[N]} \}$, $\rmdV=\{ \rmdV_{[1]} \dots \rmdV_{[N]} \}$.
\end{algorithmic}
\end{algorithm}

\subsection{Hardware-ware algorithm} Many algorithms are fast in theory, but slow in practice, due to misalignment with hardware properties \cite{Hooker2020TheHL,Saphra2023FirstTT}. 
For example, matmuls with butterfly matrices have theoretically lower complexity by using FFT, but in practice it is slow due to extensive memory  transportation operations, motivating matrices  \cite{Dao2022MonarchES,Fu2023MonarchMA} which can better align butterfly operators to GPUs. In practice it is important to reduce HBM I/O cost using techniques such as tiling and recomputation  and leverage tensor cores as much as possible. Our \textsc{FlashLinearAttention} is similar in spirit to \textsc{FlashAttention} \cite{flashattention1,flashattention2} and \textsc{FlashConvFFT} \cite{flashconvfft}, which implement I/O-aware versions of neural network layers to enable practical wallclock speedups. Concurrent work by \citet{lightning2} also proposes an I/O-aware version of linear attention, which is  similar to the non-materialization version of  \textsc{FlashLinearAttention}. We additionally propose a materialization version, which leverages sequence-level parallelism and thus allows for higher training throughput at the cost of a slightly increasing memory footprint.
\label{sec:gla_algo}

\section{Details for Chunkwise (Gated) Linear Attention}
\paragraph{Backward pass of \textsc{FlashLinearAttention}.}  
The pseduocode for backward pass of linear attention is listed in Algorithm~\ref{algo:la-chunk-bwd}.

\paragraph{Pseudo codes of GLA.}
We first present the direct adaptions of \textsc{FlashLinearAttention} to training GLA without secondary-level chunking. 
Specifically, 
Alg.~\ref{algo:gla-chunk-scan-1-fwd} and \ref{algo:gla-chunk-scan-1-bwd} shows the forward/backward pass for the materialization version; Alg.~\ref{algo:gla-chunk-scan-2-fwd} and \ref{algo:gla-chunk-scan-2-bwd} for the non-materialization version.
We show the psuedo code of our secondary-level chunking in Pytorch style in Listing~\ref{a}.
 
\definecolor{codegreen}{rgb}{0,0.6,0}
\definecolor{codegray}{rgb}{0.5,0.5,0.5}
\definecolor{codepurple}{rgb}{0.58,0,0.82}
\definecolor{backcolour}{rgb}{0.95,0.95,0.92}
 
\lstdefinestyle{mystyle}{
    backgroundcolor=\color{backcolour},   
    commentstyle=\color{codegreen},
    keywordstyle=\color{magenta},
    numberstyle=\tiny\color{codegray},
    stringstyle=\color{codepurple},
    basicstyle=\tiny,
    breakatwhitespace=false,         
    breaklines=true,                 
    captionpos=b,                    
    numbers=left,                    
    showspaces=false,                
    showstringspaces=false,
    showtabs=false,    
}
\lstset{style=mystyle}

\lstset{language=Python, basicstyle=\ttfamily\footnotesize}
\begin{lstlisting}[language=Python, caption=Pytorch-like code snippet of our two-level chunking algorithm for training GLA.  We omit the dimensions of batch size and number of heads for clarity, label=a, numbers=none]
def gated_linear_attention_forward(Q, K, V, a, C, c):
    '''
    Q/K/V: query/key/value 
    a: log forget gate 
    C/c: chunk size, subchunk size
    '''
    # L: sequence length, d: head dimension
    L, d_k = Q.shape
    d_v = V.shape[-1]
    S = torch.zeros(d_k, d_v)
    O = torch.empty_like(V)
    # cumsum of log decay within a chunk
    B = torch.empty_like(a)
    # local compute of cumulative product of decay within a chunk
    for i in range(0, L//C):
        b = torch.zeros(d_k)
        for j in range(0, C):
            b += a[i]
            B[i] = b

    for i in range(0, L // C):
        r = range(i*C,(i+1)*C) 
        # (C, d) chunking
        bq, bk, bv, bb = Q[r], K[r], V[r], B[r] 
        b = bb[-1,None]
        #inter-chunk w/ matmul
        q, k, g = bq*(bb.exp()), bk*((b-bb).exp()), b.exp()
        o = q @ S
        #hidden state update
        S = g.t() * S + k.t() @ bv 
        #intra-chunk (secondary chunking)
        for j in range(0, C // c):
            t = range(j*c, (j+1)*c)
            #(c, head_dim) subchunking
            q, k, v, b = bq[t], bk[t], bv[t], bb[t] 
            p = torch.zeros(c,c)
            #intra-subchunk w/o matmul. 
            for m in range(c):
                for n in range(m+1):
                    p[m,n]=torch.sum(q[m]*k[n]*((b[m]-b[n]).exp()))
            o[t] += p @ v
            # inter-subchunk w/ matmul    
            z = b[0, None]
            q = q * (b-z).exp()
            for u in range(0, j):
                y = range(u*c, (u+1)*c) 
                p = q @ (bk[y]*(z-bb[y]).exp()).t()
                o[t] += p@bv[y]
        O[r] = o
    return O
\end{lstlisting}

\begin{algorithm}[b!]
\footnotesize
\caption{Forward pass for gated linear attention (w. materialization)}
\label{algo:gla-chunk-scan-1-fwd}
\begin{algorithmic}
    \Require $\rmQ, \rmK, \in \R^{L \times d_k}, \rmV  \in \R^{L \times d_v}$, $\rmG =[\balpha_1 \dots \balpha_L] \in \R^{L \times d_k}$, chunk size $C$
    \State Divide $\rmQ, \rmK, \rmG$ into $N = \frac{L}{C}$ blocks $\{ \rmQ_{[1]} \dots \rmQ_{[N]} \} $, $\{ \rmK_{[1]} \dots \rmK_{[N]} \}$, $\{ \rmG_{[1]} \dots \rmG_{[N]} \} $ of size $C \times d_k$ each.
    \State Divide $\rmV$ into $N$ blocks $\{ \rmV_{[1]} \dots \rmV_{[N]} \}$ of size $C \times d_v$ each.
     \State{Initialize $\rmS = \bm{0} \in \mathbb R^{d_k\times d_v} $ on SRAM}
    \For{$n \gets 1, N$}
      \State{Write $\mathbf S$ to HBM as $\rmS_{[n]}$.} 
        \State Load $\rmK_{[n]}, \rmG_{[n]} \in \R^{C \times d_k}$ from HBM to SRAM. 
        \State Load $\rmV_{[n]} \in \R^{C \times d_v}$ from HBM to SRAM.
        \State On chip, compute $\bgamma_{[n]} \in \R^{d_k}, \bGamma_{[n]} \in \R^{C \times d_k}$ and $\tilde \rmK_{[n]} = \rmK_{[n]} \odot \bGamma_{[n]}$. %
        \State{On chip, compute $\rmS = \left(\bgamma_{[n]}^\intercal\mathbf{1}\right) \odot \mathbf{S} + \tilde \rmK_{[n]}^{\top}  \rmV_{[n]}$.}  
    \EndFor
   \ParFor{$n \gets 1, N$}
        \State Load $\rmQ_{[n]}, \rmK_{[n]}, \rmG_{[n]} \in \R^{C \times d_k}$ from HBM to SRAM.
        \State Load $\rmV_{[n]}  \in \R^{C \times d_v}$ from HBM to SRAM.
        \State Load $\rmS_{[n]} \in \R^{d_k \times d_v}$ from HBM to SRAM.
        \State On chip, construct causal mask $\rmM\in \R^{C \times C}$ %
        \State On chip, compute $\bLambda_{[n]}, \bGamma_{[n]} \in \R^{C \times d_k}$
        \State On chip, compute $\tilde \rmQ_{[n]} = \rmQ_{[n]} \odot \bLambda_{[n]}$, $\tilde \rmK_{[n]} = \rmK_{[n]} \odot \bGamma_{[n]}$, $\bar \rmK_{[n]} = \rmK_{[n]} / \bLambda_{[n]}$
        \State On chip, compute $\rmO_{[n]}^{\text{inter}} = \tilde \rmQ_{[n]} \rmS_{[n]} \in \R^{C \times d_v}$
        \State On chip, compute $\rmP = (\tilde \rmQ_{[n]} \bar \rmK^\intercal_{[n]})\odot \rmM  \in \R^{C \times C} $
        \State On chip, compute $\rmO^{\text{intra}} = \rmP \rmV_{[n]} $ 
        \State On chip, compute $ \rmO_{[n]} = \rmO^{\text{inter}} + \rmO^{\text{intra}} $
        \State Store $\rmO_{[n]}$ to HBM.
   \EndParFor
  \State \Return $\rmO = \{ \rmO_{[1]} \dots \rmO_{[N]} \}$, $\rmS=\{ \rmS_{[1]} \dots \rmS_{[N]} \}$.
\end{algorithmic}
\end{algorithm}
\begin{algorithm}[tbh!]
\footnotesize
\caption{Backward pass for gated linear attention (w. materialization)}
\label{algo:gla-chunk-scan-1-bwd}
\begin{algorithmic}
    \Require $\rmQ, \rmK, \rmG \in \R^{L \times d_k}$, $\rmV, \rmO, \rmdO \in \R^{L \times d_v}$, chunk size $C$
     \State{Initialize $\rmdS = \bm{0} \in \mathbb R^{d_k\times d_v} $ on SRAM}
    \For{$n \gets N, 1$} %
        \State Store $\rmdS$ in HBM as $\rmdS_{[n]}$
        \State{Load $\rmG_{[n]}  \in \R^{C \times d_k}$ from HBM to SRAM.}
        \State{Load $\rmQ_{[n]}  \in \R^{C \times d_k}$ from HBM to SRAM.}
        \State{Load $\rmdO_{[n]}  \in \R^{C \times d_v}$ from HBM to SRAM.}
        \State On chip, compute $\bgamma_{[n]}$, $\bGamma_{[n]}$ and $\tilde \rmQ{[n]} =  \rmQ_{[n]} \odot \bGamma{[n]}$
        \State On chip, compute $\rmdS = \left(\bgamma_{[n]}^\intercal\mathbf{1}\right) \odot \rmdS + \tilde \rmQ_{[n]}^\intercal \rmdO_{[n]}$
      \EndFor
      \ParFor{$n \gets 1, N$}
        \State{Load $\rmQ_{[n]}, \rmK_{[n]}, \rmG_{[n]} \in \R^{C \times d_k}$ from HBM to  SRAM.}
        \State Load $\rmS_{[n]} \in \R^{d_k \times d_v}$ from HBM to  SRAM.
        \State Load $\rmV_{[n]}, \rmO_{[n]}, \rmdO_{[n]} \in \R^{C \times d_v}$ from HBM to  SRAM.
        \State Load $\rmdS_{[n]} \in \R^{d_k \times d_v}$ from HBM to SRAM.
        \State On chip, construct causal mask $\rmM\in \R^{B\times B}$ %
        \State On chip, compute $\bLambda_{[n]}, \bGamma_{[n]} \in \R^{C \times d_k}$
        \State On chip, compute $\tilde \rmQ_{[n]} = \rmQ_{[n]} \odot \bLambda_{[n]}$, $\tilde \rmK_{[n]} = \rmK_{[n]} \odot \bGamma_{[n]}$, $\bar \rmK_{[n]} = \rmK_{[n]} / \bLambda_{[n]}$.
        \State On chip, compute $\rmP = (\tilde \rmQ_{[n]} \tilde \rmK^\intercal_{[n]})\odot \rmM  \in \R^{C \times C} $
        \State On chip, compute $\rmdP =( \rmdO_{[n]} \rmV_{[n]}^\intercal) \odot \rmM $
        \State On chip, compute $ \rmdbK_{[n]} = \tilde \rmQ_{[n]} \rmdP^\intercal$
        \State On chip, compute $ \rmdtK_{[n]} = \rmV_{[n]} \rmdS_{[n]}  ^\intercal$
        \State On chip, compute $\rmdK_{[n]} =  \rmdtK_{[n]} \odot \bGamma_{[n]}  + \rmdbK_{[n]} / \bLambda_{[n]} $
        \State{On chip, compute $\rmdtQ_{[n]} = \rmdP \, \bar \rmK_{[n]} + \rmdO_{[n]} \rmS_{[n]}^\intercal $}
        \State On chip, compute $\rmdQ_{[n]} =  \rmdtQ_{[n]} \odot \bLambda_{[n]} $
        \State On chip, compute $\rmdV_{[n]} = \rmP^\intercal \, \rmdO_{[n]} + \tilde \rmK_{[n]} \rmdS_{[n]} $
        \State Store $\rmdK_{[n]}, \rmdV_{[n]}$ in HBM.
      \EndParFor
  \State Let $\rmdQ = \{ \rmdQ_{[1]} \dots \rmdQ_{[N]} \}$, $\rmdK=\{ \rmdK_{[1]} \dots \rmdK_{[N]} \}$, $\rmdV=\{ \rmdV_{[1]} \dots \rmdV_{[N]} \}$.
  \State Compute $\rmdA = \rmQ \odot \rmdQ - \rmK \odot \rmdK $, $\rmdG = \revcum(\rmdA) $
  \State \Return $\rmdQ, \rmdK, \rmdV, \rmdG$
\end{algorithmic}
\end{algorithm}

\begin{algorithm}[b!]
\small
\caption{Forward pass for gated linear attention (w/o. materialization)}
\label{algo:gla-chunk-scan-2-fwd}
\begin{algorithmic}
    \Require $\rmQ, \rmK, \in \R^{L \times d_k}, \rmV  \in \R^{L \times d_v}$, $\rmG =[\balpha_1 \dots \balpha_L] \in \R^{L \times d_k}$, chunk size $C$
    \State Divide $\rmQ, \rmK, \rmG$ into $N = \frac{L}{B}$ blocks $\{ \rmQ_{[1]} \dots \rmQ_{[N]} \} $, $\{ \rmK_{[1]} \dots \rmK_{[N]} \}$, $\{ \rmG_{[1]} \dots \rmG_{[N]} \} $ of size $C \times d_k$ each.
    \State Divide $\rmV$ into $N$ blocks $\{ \rmV_{[1]} \dots \rmV_{[N]} \}$ of size $C \times d_v$ each.
     \State{Initialize $\rmS = \bm{0} \in \mathbb R^{d_k\times d_v} $ on SRAM}
    \For{$n \gets 1, N$}
      \State{Write $\mathbf S$ to HBM as $\rmS_{[n]}$.} 
        \State Load $\rmQ_{[n]}, \rmK_{[n]}, \rmG_{[n]} \in \R^{C \times d_k}$ from HBM to SRAM.
        \State Load $\rmV_{[n]}  \in \R^{C \times d_v}$ from HBM to SRAM.
        \State On chip, compute $\bgamma_{[n]} \in \R^{d_k}, \bGamma_{[n]} \in \R^{C \times d_k}$ and $\tilde \rmK_{[n]} = \rmK_{[n]} \odot \bGamma_{[n]}$. %
        \State On chip, construct causal mask $\rmM\in \R^{C\times C}$ %
        \State On chip, compute $\bLambda_{[n]}, \bGamma_{[n]} \in \R^{C \times d_k}$
        \State On chip, compute $\tilde \rmQ_{[n]} = \rmQ_{[n]} \odot \bLambda_{[n]}$, $\tilde \rmK_{[n]} = \rmK_{[n]} \odot \bGamma_{[n]}$, $\bar \rmK_{[n]} = \rmK_{[n]} / \bLambda_{[n]}$.
        \State On chip, compute $\rmO_{[n]}^{\text{inter}} = \tilde \rmQ_{[n]} \rmS_{[n]} \in \R^{C \times d_v}$
        \State On chip, compute $\rmP = (\tilde \rmQ_{[n]} \bar \rmK^\intercal_{[n]})\odot \rmM  \in \R^{C \times C} $
        \State On chip, compute $\rmO^{\text{intra}} = \rmP \rmV_{[n]} $
        \State On chip, compute $ \rmO_{[n]} = \rmO^{\text{inter}} + \rmO^{\text{intra}} $
        \State Store $\rmO_{[n]}$ to HBM.
        \State{On chip, compute $\rmS = \left(\bgamma_{[n]}^\intercal \mathbf{1}\right) \odot \mathbf{S} + \tilde \rmK_{[n]}^{\top}  \rmV_{[n]}$.}  
   \EndFor
  \State \Return $\rmO = \{ \rmO_{[1]} \dots \rmO_{[N]} \}$.
\end{algorithmic}
\end{algorithm}

\begin{algorithm}[tbh!]
\small
\caption{Backward pass for gated linear attention (w/o. materialization)}
\label{algo:gla-chunk-scan-2-bwd}
\begin{algorithmic}
    \Require $\rmQ, \rmK, \rmG \in \R^{L \times d_k}$, $\rmV, \rmO, \rmdO \in \R^{L \times d_v}$, chunk size $C$
    \State{Initialize $\rmS = \bm{0} \in \R^{d_k\times d_v} $ on SRAM}
    \For{$n \gets 1, N$} %
        \State{Load $\rmG_{[n]}  \in \R^{C \times d_k}$ from HBM to SRAM.}
        \State{Load $\rmQ_{[n]}  \in \R^{C \times d_k}$ from HBM to SRAM.}
        \State{Load $\rmdO_{[n]}  \in \R^{C \times d_v}$ from HBM to SRAM.}
        \State On chip, compute $\bgamma_{[n]} \in \R^{d_k}, \bGamma_{[n]} \in \R^{C \times d_k}$ and $\tilde \rmK_{[n]} = \rmK_{[n]} \odot \bGamma_{[n]}$.
        \State On chip, compute $\rmdP = \rmdO_{[n]} \rmV_{[n]}^\intercal $
        \State{On chip, compute $\rmdtQ_{[n]} = \rmdP \, \tilde \rmK_{[n]} + \rmdO_{[n]} \rmS^\intercal $}
        \State On chip, compute $\rmdQ= \rmdtQ_{[n]} \odot \bGamma{[n]} $
        \State Store $\rmdQ_{[n]}$ to HBM.
        \State{On chip, compute $\rmS = \left(\bgamma_{[n]}^\intercal\mathbf{1}\right) \odot \mathbf{S} + \tilde \rmK_{[n]}^{\top}  \rmV_{[n]}$.}  
      \EndFor
    \State{Initialize $\rmdS = \bm{0} \in \mathbb R^{d_k\times d_v} $ on SRAM}
      \For{$n \gets N, 1$}
        \State{Load $\rmQ_{[n]}, \rmK_{[n]}, \rmG_{[n]} \in \R^{C \times d_k}$ from HBM to  SRAM.}
        \State Load $\rmV_{[n]}, \rmO_{[n]}, \rmdO_{[n]} \in \R^{C \times d_v}$ from HBM to  SRAM.
        \State On chip, construct causal mask $\rmM\in \R^{C\times C}$ %
        \State On chip, compute $\bLambda_{[n]}, \bGamma_{[n]} \in \R^{C \times d_k}$
        \State On chip, compute $\tilde \rmQ_{[n]} = \rmQ_{[n]} \odot \bLambda_{[n]}$, $\tilde \rmK_{[n]} = \rmK_{[n]} \odot \bGamma_{[n]}$.
        \State On chip, compute $\rmP = (\tilde \rmQ_{[n]} \tilde \rmK^\intercal_{[n]})\odot \rmM  \in \R^{C \times C} $
        \State On chip, compute $\rmdP = (\rmdO_{[n]} \rmV_{[n]}^\intercal) \odot \rmM $
                \State On chip, compute $ \rmdbK_{[n]} = \tilde \rmQ_{[n]} \rmdP^\intercal$
        \State On chip, compute $ \rmdtK_{[n]} = \rmV_{[n]} \rmdS_{[n]}  ^\intercal$
        \State On chip, compute $\rmdK_{[n]} =  \rmdtK_{[n]} \odot \bGamma_{[n]}  + \rmdbK_{[n]} / \bLambda_{[n]} $
        \State On chip, compute $\rmdV_{[n]} = \rmP^\intercal \, \rmdO_{[n]} + \tilde \rmK_{[n]} \rmdS $
        \State Store $\rmdQ_{[n]}, \rmdK_{[n]}, \rmdV_{[n]}$ in HBM.
        \State On chip, compute $\rmdS = \left(\bgamma_{[n]}^\intercal \mathbf{1} \right) \odot \rmdS + \tilde \rmQ_{[n]}^\intercal \rmdO_{[n]}$
      \EndFor
  \State Let $\rmdQ = \{ \rmdQ_{[1]} \dots \rmdQ_{[N]} \}$, $\rmdK=\{ \rmdK_{[1]} \dots \rmdK_{[N]} \}$, $\rmdV=\{ \rmdV_{[1]} \dots \rmdV_{[N]} \}$.
  \State Compute $\rmdA = \rmQ \odot \rmdQ - \rmK \odot \rmdK $, $\rmdG = \revcum(\rmdA) $
  \State \Return $\rmdQ, \rmdK, \rmdV, \rmdG$
\end{algorithmic}
\end{algorithm}

\paragraph{Derivations of $\dblogalpha_t$.}

We show the derivations for the following gradient form.
\begin{align*}
    \vdlogb_t &= \vk_t \odot \vdk_t - \vq_t \odot \vdq_t, \\
   \dblogalpha_t  &= \sum_{t \leq i \leq L} \vdlogb_i.
\end{align*}

By unrolling the recurrence, we have
\begin{align*}
\vo_t =  \vq_t \rmS_t  &= \sum_{i=1}^{t}(\vq_t \odot \vb_t) \left(\frac{\vk_i}{\vb_i}\right)^{\top} \vv_i \\
&=  \sum_{i=1}^{t}(\vq_t \odot \exp(\log \vb_t)) \left(\vk_i \odot {\exp(- \log \vb_i)}\right)^{\top} \vv_i
\end{align*}
where at the second step, we apply a trivial identity: $\exp(\log x) = x$.
We first derive the gradients wrt. query/key vectors,
\begin{align*}
& \vdq_t = \sum_{ i = 1}^{t} \langle \vdo_t, \vv_i\rangle  \vb_t \odot \vk_i / \vb_i,  \\
& \vdk_i = \sum_{t = i}^{L} \langle \vdo_t, \vv_i\rangle  \vq_t \odot  \vb_t  / \vb_i.
\end{align*}
Then for the gradients wrt. the logits of the accumulative gates,
\begin{align*}
& \vdlogb_t  = \vq_t \odot \underbrace{\sum_{i=1}^{t} \langle \vdo_t, \vv_i\rangle  \odot  \vb_t  \odot  \vk_i / \vb_i}_{  \vdq_t} - \vk_t \odot \underbrace{\sum_{i = t}^{L} \langle \vdo_i, \vv_t\rangle  \vq_i \odot  \vb_i  / \vb_t}_{\vdk_t}.
\end{align*}
where we change the index notation for the $\vdk$ term.
It now becomes clear that 
\begin{align*}
& \vdlogb_t =  \vq_t \odot \vdq_t  - \vk_t \odot \vdk_t.
\end{align*}
Since $\vlogb_t = \sum_{i=1}^t \vlogalpha_i$, we get $\dblogalpha_t  = \sum_{t = i}^L \vdlogb_i$.

\section{General Gated Linear Attention}
\label{sec:ggla}

In the main paper, we use a simplified parameterization where $\bbeta$ is fixed to $\mathbf{1}$ in the following gated linear attention.
\begin{align*}
\rmS_t = (\balpha_t^{\top} \bbeta_t) \odot \rmS_{t-1} + \vk_t^{\top} \vv_t,
\end{align*}
Though empirically we found that making $\bbeta$ learnable does not lead to performance gain, we show here that the general form still enjoys parallel form and chunk-wise form, which could be potentially useful for future development of linear attention models.

\subsection{Parallel form}

By unrolling the recurrence we have,
\begin{align}
\vo_t = \vq_t \rmS_t 
= \vq_t \sum_{i=1}^{t} \big((\prod_{i+1}^{t} \rmG_{i}) \odot (\vk_i^{\intercal}\vv_i)\big) 
\end{align}

By taking advantage of the mixed product property of Kronercker/outer product, we have
\begin{align}
(\prod_{j=i+1}^{t} \rmG_j) \odot (\vk_i^{\intercal} \vv_i) &= \big((\frac{\vb_t}{\vb_i})^{\intercal}(\frac{\vd_t}{\vd_i})\big) \odot (\vk_i^{\intercal} \vv_i) \\
&=\left( \frac{\vb_t}{\vb_i}\odot \vk_i \right)^{\intercal} \left( \frac{\vd_t}{\vd_i} \odot \vv_i \right) 
\end{align}
where $\vb_t = \prod_{j=1}^t \balpha_j, \vd_t = \prod_{j=1}^t \bbeta_j$.
By plugging it into the expanded recurrence, we have the following form. 
\begin{align}
\vo_t = \vq_t \rmS_t 
&= 
\vq_t \sum_{i=1}^{t} \big((\prod_{i+1}^{t} \rmG_{i}) \odot (\vk_i^{\intercal}\vv_i)\big) \\
    &= \vq_t \sum_{i=1}^{t} \left( \frac{\vb_t}{\vb_i}\odot \vk_i \right)^{\intercal} \left( \frac{\vd_t}{\rmB_i} \odot \vv_i \right) \\ 
    &= \sum_{i=1}^{t} \left(\vq_t \left( \frac{\vb_t}{\vb_i}\odot \vk_i \right)^{\intercal}\right) \left( \frac{\vd_t}{\vd_i} \odot \vv_i \right) \label{eq:gla_parallel_1} \\
    &= \sum_{i=1}^{t} \underbrace{\left\langle \vq_t, \frac{\vb_t}{\vb_i} \odot \vk_t \right\rangle}_{\R^{1\times 1}} \underbrace{\left( \frac{\vd_t}{\vd_i} \odot \vv_t \right)}_{\mathbb{R}^{1\times d_v}}  \\    
    &= \sum_{i=1}^{t} \left(\left\langle \vq_t \odot \vb_t, \frac{\vk_i}{\vb_i}\right\rangle \frac{\vv_i}{\vd_i} \right) \odot \vd_t \qquad  \label{eq:gla_parallel_2} \\
    &= \sum_{i=1}^{t} \left( \left(\vq_t \odot \vb_t\right)\left(\frac{\vk_i}{\vb_i}\right)^{\intercal} \left(
    \frac{\vv_i}{\vd_i} \right) \right) \odot \vd_t \quad \in \mathbb{R}^{1\times d_v} 
\end{align}
Eq.~\ref{eq:gla_parallel_1} is by linearity and associative property of matrix multiplication, Eq.~\ref{eq:gla_parallel_2} is derived based on $\langle \va, \vb \odot \vc \rangle = \langle \va \odot \vb, \vc \rangle$. The final form has following equivalent parallel form similar to the parallel form of linear/softmax attention.
\begin{align}
    \tilde \rmQ &= \rmQ \odot \rmB \qquad \tilde \rmK = \rmK / \rmB \qquad \tilde \rmV = \rmV / \rmD \\
    \tilde \rmO &= (\tilde \rmQ \tilde \rmK^\intercal \odot \rmM) \tilde \rmV \qquad \rmO = \tilde \rmO \odot \rmD
    \label{eq:gla_QKV}
\end{align}
where $\rmQ, \rmK, \rmB \in \R^{L \times d_k}$,
 $\rmV, \rmD \in \R^{L \times d_v}$,
 $\rmM \in \R^{L \times L}$ denotes the causal mask.

\subsection{Chunkwise parallel form}

Now we show that the chunkwise parallel form for efficient training of general linear attention. Suppose $\rmX$ is now split into $\frac{L}{C}$ chunks, each of length $C$. Let $\rmS_{[i]} \in \R^{d_k \times d_v}$ be the chunk-level hidden state after processing $i$ chunks, i.e., $\rmS_{[i]}:=\rmS_{iC}$. Further let $\rmK_{[i+1]}:=\rmK_{iC+1:(i+1)C} \in \mathbb{R}^{C\times d_k}$, $\rmV_{[i+1]}:=\rmV_{iC+1:(i+1)C} \in \mathbb{R}^{C\times d_v}$. The inter-chunk recurrence is then given by,
\begin{align*}
\rmS_{[i+1]} 
&= \left(\Big(\frac{\rmB_{(i+1)C}}{\rmB_{iC}}\Big)^{\intercal}\Big(\frac{\rmD_{(i+1)C}}{\rmD_{iC}}\Big) \right) \odot \rmS_{[i]} +  \big(\rmB^{\prime}_{[i+1]} \odot \rmK_{[i+1]}\big)^{\intercal} \big(\rmD^{\prime}_{[i+1]} \odot \rmV_{[i+1]}\big),
\label{eq:gla_inter_chunk_recur}
\end{align*}
 where $(\rmB^{\prime}_{[i+1]})_j = \frac{\rmB_{(i+1)C}}{\rmB_{iC+j}} \in \mathbb{R}^{1\times d_k}$ and $(\rmD^{\prime}_{[i+1]})_j = \frac{\rmD_{(i+1)C}}{\rmD_{iC+j}} \in \mathbb{R}^{1\times d_v}$ for $j\in [1, C]$, $i \in [0, L/C-1]$. (Therefore we have $\rmB^{\prime}_{[i+1]} \in \R^{C \times d_k}, \rmD^{\prime}_{[i+1]} \in \R^{C \times d_v}$.) The intra-chunk parallel computation is then given by,
\begin{align}
\tilde \rmO_{[i+1]} &= \underbrace{\left((\rmQ_{[i+1]} \odot \rmB^{\dagger}_{[i+1]}) \rmS_{[i]} \right) \odot \rmD^{\dagger}_{[i+1]}}_{\text{inter-chunk}} + \underbrace{
(\tilde\rmQ_{[i+1]} \tilde\rmK_{[i+1]}^{\intercal} \odot \rmM)\tilde\rmV_{[i+1]}
}_{\text{intra-chunk}}, \\ &\rmO_{[i+1]}  = \tilde \rmO_{[i+1]} / \rmD^{\dagger}_{[i+1]},
\label{eq:gla_inter_intra} 
\end{align}
 where $(\rmB_{[i+1]}^{\dagger})_j = \frac{\rmB_{iC+j}}{\rmB_{iC}} \in \mathbb{R}^{1\times d_k}$ and $(\rmD_{[i+1]}^{\dagger})_j = \frac{\rmD_{iC+j}}{\rmD_{iC}} \in \mathbb{R}^{1\times d_v}$  for $j\in [1, C]$, $i \in [0, L/C-1]$. 
 Subsequently, we have $\tilde\rmQ_{[i+1]}=\rmQ_{[i+1]} \odot \rmB_{[i+1]}^{\dagger}, \tilde\rmK_{[i+1]}=\frac{\rmK_{[i+1]}}{\rmB_{[i+1]}^{\dagger}}, \tilde\rmV_{[i+1]}=\rmV_{[i+1]} \odot \rmD_{[i+1]}^{\dagger} $. For initial values, we set $\rmS_0 = \vzero$, $\rmB_0=\mathbf{1}$, $\rmD_0=\mathbf{1}$. Intuitively, $\rmB^{\prime}_{[i]} $ encodes the cumulative decay from the start of a chunk which will be used to propagate the hidden states from the previous chunk $\rmS_{[i]}$;   $\rmB^{\dagger}_{[i]}$ encodes the decay to the end of a chunk which will be used to accumulate information to be added to the next hidden state $\rmS_{[i+1]}$.

The chunkwise form given here is a generalization of several existing forms for linear attention. If we set $\rmA_{ij}=1$, $\rmB_{ij} = 1$, it reduces to the chunk-wise form presented in the main paper for vanilla linear attention; if we set $\rmA_{ij}=1$, $\rmB_{ij} = \gamma^{i+1}$, it becomes RetNet's chunk-wise form \citep{sun2023retentive}. As such, our formulation can be regarded as a generalized chunk-wise parallel form for linear attention that enables fine-grained data-dependent decay. 
\paragraph{Memory-efficient computation of $\dbalpha$ and $\dbbeta$}

In the general form, we show that the gradient wrt. $\balpha$  and $\bbeta$ admits the following closed form, which allows computing $\dbalpha$ and $\dbbeta$ without instantiating $\rmS$ in HBM.
\begin{align*}
    \vdlogb_t &= \vk_t \odot \vdk_t - \vq_t \odot \vdq_t, \\
   \dblogalpha_t  &= \sum_{t \leq i \leq L} \vdlogb_i \\
   \vdlogd_t &= \vo_t \odot \vdo_t - \vv_t \odot \vdv_t, \\
   \dblogbeta_t  &= \sum_{t \leq i \leq L} \vdlogd_i.
\end{align*}
where  $\vlogb_t = \sum_{i=1}^t \log \balpha_i$, $\vlogd_t = \sum_{i=1}^t \bbeta_i$ (or alternatively $\vb_t = \prod_{i=1}^t \balpha_i$, $\vd_t = \prod_{i=1}^t \bbeta_i$).  We apply the trick to compute $\vdlogb_t$ and $\vdlogd_t$ for the following cumulative-sum form.

\begin{align*}
    \vo_t = \sum_{i=1}^{t} \left( \left(\vq_t \odot \vb_t\right)\left(\frac{\vk_i}{\vb_i}\right)^{\intercal} \left(
    \frac{\vv_i}{\vd_i} \right) \right) \odot \vd_t \quad \in \mathbb{R}^{1\times d_v}.
\end{align*}
The gradient of $\vlogb_t$ comes from two sources: one associated with $\vq_t$, the other associated with $\vk_i$. Similarly, $\vlogd_t$ comes from both $\vo_t$ and $\vv_i$.
\begin{align*}
\vdlogb_t  &= \vq_t \odot \underbrace{\sum_{i=1}^{t} \langle \vdo_t, \frac{\vd_t}{\vd_i} \vv_i \rangle  \odot  \vb_t  \odot  \vk_i / \vb_i}_{  \vdq_t} - \vk_t \odot \underbrace{\sum_{i = t}^{L} \langle \vdo_i, \frac{\vd_i}{\vd_t} \vv_t \rangle  \vq_i \odot  \vb_i  / \vb_t}_{\vdk_t} \\
\vdlogd_t  &= \vdo_t \odot \underbrace{\sum_{i=1}^{t} \left( \left(\vq_t \odot \vb_t\right)\left(\frac{\vk_i}{\vb_i}\right)^{\intercal} \left(
    \frac{\vv_i}{\vd_i} \right) \right) \odot \vd_t}_{\vo_t} - \vv_t \odot \underbrace{\sum_{i=t}^{L} \left( \left(\vq_i \odot \vb_i\right)\left(\frac{\vk_t}{\vb_t}\right)^{\intercal} \left(
    \frac{1}{\vd_t} \right) \right) \odot \vd_i}_{\vdv_t} 
\end{align*}
The trick applied there is that $\frac{\partial f(\va \odot \vb )}{\partial \vlogb} = \va \odot \frac{\partial f(\va \odot \vb )}{\partial \va}$ and  $\frac{\partial f(\va / \vb )}{\partial \vlogb}  = - \frac{\partial f(\va / \vb )}{\partial \va} \odot \va $.

\vspace{-2mm}
\section{Additional Experimental Results}
\label{app:d}
\begin{table}[ht]
\centering
\tiny
\begin{tabular}{l|ccccccccccccc}
\toprule
\textbf{Model}  & \textbf{Wiki.}  &  \textbf{LMB.} & \textbf{LMB.} & \textbf{PIQA} &    \textbf{Hella.} & \textbf{Wino.} & \textbf{ARC-e} &  \textbf{ARC-c} & \textbf{COPA} &  \textbf{OBQA} & \textbf{SciQA} &  \textbf{BoolQ} &   \textbf{Avg.}  \\
 & ppl $\downarrow$  & ppl $\downarrow$  & acc $\uparrow$  & acc $\uparrow$ &   acc\_norm $\uparrow$  & acc $\uparrow$  & acc $\uparrow$ & acc\_norm $\uparrow$  & acc $\uparrow$ & acc\_norm $\uparrow$ & acc $\uparrow$ & acc $\uparrow$ \\
\midrule
\small{\textit{0-shot}}  \\
Transformer++ 340M & 28.39 & 	42.69	& 31.0 & 	63.3 &	34.0 &	50.4 & 44.5	& 24.2 & 66.0 &	28.4	& 73.8	& 60.9 & 47.7 \\
RetNet 350M & 32.33 & 49.19	& 28.6 &	63.5 &	33.5 & 52.5 &	44.5	& 23.4 &	63 &	28.4 & 73.1	& 60.0 & 47.1 \\
Mamba 350M & 28.39 &  39.66 & 30.6 & 65.0 & 35.4 & 50.1 & 46.3 & 23.6 & 71.0 & 28.4 & 73.7 & 52.6 & 47.7 \\
GLA-Transformer 340M & 28.65 & 43.35 & 30.3 & 64.8 & 34.5 &	51.4 & 45.1	& 22.7 & 70.0 &29.2 & 73.2 & 58.7 & 48.0 \\
\midrule 
\small{\textit{0-shot}} \\
Transformer++ 1.3B & 16.85 & 13.44 &  48.9 & 70.8 & 49.6 & 53.6 & 56.0 & 26.5 & 
 75.0 & 29.8 & 83.6 & 52.3 & 54.6 \\
RetNet 1.3B & 18.64  & 17.27 & 43.3 & 70.0 & 47.3 & 52.5 & 54.8 & 25.6 & 70.0 & 31.4 & 82.3 & 57.1 & 53.4  \\
Mamba 1.3B &  17.06  & 13.89 & 46.2 & 72.2 &  40.1 & 54.1 &  59.0 &	28.2 & 74.0 & 33.0 & 83.1 & 59.1 & 54.9     \\
GLA-Transformer 1.3B & 17.22 & 14.47 & 46.9 & 71.8 & 49.8 & 53.9 & 57.2 & 26.6 & 73.0 & 32.4 & 84.7 & 58.5 & 55.5 \\
\midrule
\small{\textit{5-shot}} \\
Transformer++ 1.3B & - & 16.80 & 42.9 & 70.2 & 50.3 &	53.8 & 60.5	& 28.7 & 75.0 & 33.8 & 90.7 & 46.0 & 55.2 \\
RetNet 1.3B & - &  23.27 & 37.3 & 69.8 & 47.5 & 51.1 & 58.5 &	27.4 & 72.0 & 31.8 & 87.5 & 45.3 & 52.8 \\
Mamba 1.3B & - & 23.00 & 31.4 & 71.4 &  51.2	 & 54.1 & 60.1 & 30.4  & 79.0 & 33.8 & 88.5 & 47.7 & 55.4 \\
GLA-Transformer 1.3B & - & 18.87 & 41.1 & 71.9 & 49.9 & 54.4 & 61.8 & 28.4 & 75.0 & 34.2 & 90.4 & 56.9 & 56.4 \\
\bottomrule
\end{tabular}
\centering
\caption{Extended zero- and five-shot performance results. All models are trained on the same subset of SlimPajama dataset with Mistral tokenizer. The 340M/1.3B models are trained for 15B/100B tokens respectively. The last column shows the average of all accuracies.}
\label{tab:extended_results}
\end{table}

The complete results on all 11 tasks, including the 5-shot results for the 1.3B models, are shown in Table~\ref{tab:extended_results}.

\end{document}